%% file: main.tex
\documentclass[10pt,journal,compsoc]{IEEEtran}
\usepackage[utf8]{inputenc}
\usepackage{amsmath,amssymb,amsfonts,amsthm,bm}
\usepackage{algorithm}
\usepackage{algorithmicx}
\usepackage{algpseudocode}

\algnewcommand\algorithmicinput{\textbf{Input:}}
\algnewcommand\algorithmicoutput{\textbf{Output:}}
\algnewcommand\INPUT{\item[\algorithmicinput]} 
\algnewcommand\OUTPUT{\item[\algorithmicoutput]} 

\usepackage{array}
\usepackage{textcomp}
\usepackage{stfloats}
\usepackage{url}
\usepackage{verbatim}
\usepackage{graphicx}

\usepackage{color}
\usepackage[acronyms]{glossaries}

\usepackage{longtable}
\usepackage{multirow}
\usepackage{tabularx}
\usepackage{booktabs}
\usepackage{array}

\usepackage{graphicx}
\usepackage{subcaption}

\usepackage{pgfplots}
\pgfplotsset{compat=newest}
\usepackage{tikz}
\usetikzlibrary{calc}
\usetikzlibrary{fit}
\usetikzlibrary{positioning}
\usetikzlibrary{spy}
\usetikzlibrary{arrows.meta}

\usepackage{enumitem}
\DeclareMathOperator*{\argmin}{arg\,min}
\newcommand{\abs}[1]{\lvert #1 \rvert}
\newcommand{\round}[1]{\lfloor #1 \rceil}
\newcommand{\floor}[1]{\lfloor #1 \rfloor}

\usepackage[hidelinks]{hyperref}

\newcommand{\transpose}{\text{T}}

\ifCLASSOPTIONcompsoc
  \usepackage[nocompress]{cite}
\else
  \usepackage{cite}
\fi

\newacronym{sbl}{SBL}{sparse Bayesian learning}
\newacronym{ml}{ML}{maximum likelihood}
\newacronym{pdf}{PDF}{probability density function}
\newacronym{snr}{SNR}{signal-to-noise ratio}
\newacronym{em}{EM}{expectation-maximization}
\newacronym{awgn}{AWGN}{additive white Gaussian noise}
\newacronym{nmse}{NMSE}{normalized mean squared error}
\newacronym{mmse}{MMSE} {minimum mean squared error}
\newacronym{elbo}{ELBO}{evidence lower bound}
\newacronym{kl}{KL}{Kullback-Leibler}

\begin{document}
    \title{\huge MB-RACS: Measurement-Bounds-based Rate-Adaptive Image Compressed Sensing Network}

    \author{Yujun Huang, Bin Chen,~\textit{Member, IEEE}, Naiqi Li, Baoyi An, Shu-Tao Xia,~\textit{Member, IEEE},\\ and Yaowei Wang\vspace*{-5mm} 
    \IEEEcompsocitemizethanks{
    
    \IEEEcompsocthanksitem Y. Huang, N. Li and S.-T. Xia are with Tsinghua Shenzhen International Graduate School, Tsinghua University, Shenzhen, Guangdong, 518055 China, and also with Research Center of Artificial Intelligence, Pengcheng Laboratory, Shenzhen, Guangdong 518055, China. (e-mail: huangyj20@mails.tsinghua.edu.cn, linaiqi@sz.tsinghua.edu.cn, xiast@sz.tsinghua.edu.cn.)
    \IEEEcompsocthanksitem B. Chen is with Harbin Institute of Technology, Shenzhen, Guangdong, 518055 China, and also with the Research Center of Artificial Intelligence, Pengcheng Laboratory, Shenzhen, Guangdong 518055, China. (e-mail: chenbin2021@hit.edu.cn.) (\emph{Corresponding author: Bin Chen.})
    \IEEEcompsocthanksitem B. An is Huawei Technologies Company Ltd., Shenzhen, Guangdong 518055, China (e-mail: anbaoyi@huawei.com.)
    \IEEEcompsocthanksitem Y. Wang is with Pengcheng Laboratory, Shenzhen, Guangdong, 518055 China. (e-mail: wangyw@pcl.ac.cn)
    }
             
    }
    \markboth{IEEE TRANSACTIONS ON PATTERN ANALYSIS AND MACHINE INTELLIGENCE}%
    {Shell \MakeLowercase{\textit{et al.}}: Bare Demo of IEEEtran.cls for Computer Society Journals}
    
    \input{Section/abstract}
    \maketitle
    \IEEEdisplaynontitleabstractindextext
    \IEEEpeerreviewmaketitle
    \glsresetall
	
    \input{Section/introduction}	
    \input{Section/related_work}
    \input{Section/proposed_method}

    \input{Section/experiments_and_results_section/experiments_and_results}

    \input{Section/conclusions}
    \bibliographystyle{IEEEtran}
    \bibliography{Reference/reference_CS}
    \input{Section/biography}

\end{document}


\input{Section/appendix}

%% file: Section/abstract.tex
\IEEEtitleabstractindextext{
\begin{abstract}
Conventional compressed sensing (CS) algorithms typically apply a uniform sampling rate to different image blocks. A more strategic approach could be to allocate the number of measurements adaptively, based on each image block's complexity. In this paper, we propose a Measurement-Bounds-based Rate-Adaptive Image Compressed Sensing Network (MB-RACS) framework, which aims to adaptively determine the sampling rate for each image block in accordance with traditional measurement bounds theory. Moreover, since in real-world scenarios statistical information about the original image cannot be directly obtained, we suggest a multi-stage rate-adaptive sampling strategy. This strategy sequentially adjusts the sampling ratio allocation based on the information gathered from previous samplings. We formulate the multi-stage rate-adaptive sampling as a convex optimization problem and address it using a combination of Newton's method and binary search techniques. Additionally, we enhance our decoding process by incorporating skip connections between successive iterations to facilitate a richer transmission of feature information across iterations. Our experiments demonstrate that the proposed MB-RACS method surpasses current leading methods, with experimental evidence also underscoring the effectiveness of each module within our proposed framework.
\end{abstract}
}

%% file: Section/introduction.tex
\section{Introduction}\label{sec:introduction}
Compressed sensing (CS) \cite{1614066, candes2006stable, 1542412} has established a novel framework for sampling and reconstruction with theoretical guarantees, which permits the remarkably precise reconstruction of sparse signals with considerably fewer measurements than that stipulated by the Nyquist sampling theorem.
This mathematical theorem posits that, given a randomly formed measurement matrix $\boldsymbol{\Phi} \in \mathbb{R}^{m\times n}$, for any sparse signal $\textbf{\textit{x}}\in\mathbb{R}^n$, provided the number of measurements $m\geq Ck\log(n/k)$, the original signal can be accurately reconstructed from the sampled signal $\textbf{\textit{y}}=\boldsymbol{\Phi} \textbf{\textit{x}}$. Here, $C$ is a constant, and $k=\|\Psi \textbf{\textit{x}}\|_0$ denotes the count of non-zero values in $\textbf{\textit{x}}$'s sparse representation after linear transformation, also termed the sparsity of $\textbf{\textit{x}}$. For the signals encountered in practice, their $k$ is often small, and therefore the number of required measurements will be small as well. This impressive result has fueled the extensive application of CS across numerous disciplines, encompassing magnetic resonance imaging (MRI) \cite{4472246, https://doi.org/10.1002/mrm.21391}, computed tomography \cite{8271999}, wireless broadcast \cite{yin2016compressive, 6457441}, network measurement \cite{265057}, among others.

Classical image CS approaches, termed optimization-based methods, frame the ill-posed linear inverse problem of CS reconstruction as an optimization problem based on image priors. The solution of these problems are then approximated using iterative methods. Notably, numerous papers have introduced reconstruction algorithms premised on diverse priors such as sparsity \cite{5414429}, local smoothness \cite{li2013efficient}, non-local self-similarity \cite{6341094}, and group-based sparsity \cite{6814320}. Nonetheless, the priors deployed in optimization-based methods do not wholly align with data traits, and the complexity required for iterative solutions is typically high.

In recent years, numerous deep image CS methods have emerged, merging deep learning concepts with CS mechanisms, leading to improved reconstruction quality and speed. \cite{8765626} designed an end-to-end deep image CS structure, employing a data-driven strategy to guide the learning of measurement matrices and the reconstruction network. \cite{Zhang_2018_CVPR, 9298950,9019857} proposed the deep unfolding method that converts the non-differentiable optimization method in traditional CS reconstruction into an iterative network structure. Integrating optimization-based methods with data-driven understanding led to a significant improvement in algorithm performance.

In addition to the learning of the measurement matrix and the design of the reconstruction network, there exists a range of research concerned with aspects such as adaptive sampling ratio allocation and model scalability. \cite{5585813} proposes a compressed sampling scheme based on saliency information. By considering human visual attention, more perceptual resources are allocated to salient regions, thereby improving image reconstruction quality. \cite{9159912} utilizes a similar CS ratio allocation method and puts forward a multi-channel deep network structure. \cite{9854112} puts forward a content-aware scalable deep CS network dubbed CASNet. It employs a data-driven saliency detector to generate saliency maps for adaptive sampling ratio allocation. \cite{Shi_2019_CVPR} and \cite{9467810} respectively explored the scalability of the count of measurements and the scalability for arbitrary sampling matrices.


Although saliency-based rate-adaptive sampling has achieved high performance, it only suggests which areas need more measurements without a clear justification for the exact number required. In this paper, we introduce the \textbf{M}easurement-\textbf{B}ounds-based \textbf{R}ate-\textbf{A}daptive Image \textbf{C}ompressed \textbf{S}ensing Network (MB-RACS), a method rooted in the measurement bounds theory of CS that determines the number of measurements a signal requires. We first propose a single-stage measurement-bounds-based rate-adaptive sampling method that calculates the measurement bounds based on block-wise sparsity information and then allocates the sampling rate based on these bounds. In some practical applications where the sparsity information of the original image cannot be accessed before sampling, we suggest a multi-stage rate-adaptive sampling strategy. Under this strategy, the sampling process is broken down into several phases. Data from initial phases guides our prediction of measurement bounds, which further informs the allocation of sampling rates for the following phases. However, complete adherence to predicted measurement bounds in assigning sampling rates is not feasible because some of the sampling rates have already been allocated in previous sampling stages. Consequently, we present a convex-optimization-based algorithm to optimize the allocation of sampling rates. Utilizing both Newton's method and the bisection method, we identify the optimal solution for this optimization problem. Moreover, to ensure thorough propagation of feature information during the forward phase, we incorporate cross-iteration skip connections within the decoder.

The contributions of this paper are as follows:

\begin{itemize}[leftmargin=1em]
\item We propose MB-RACS, a measurement-bounds-based rate-adaptive image CS method, wherein each block's sampling rate is proportionally aligned with its respective measurement bounds. Based on this, a single-stage measurement-bounds-based rate-adaptive sampling method is implemented.

\item A multi-stage rate-adaptive sampling approach is presented for some applications where the image's sparsity cannot be accessed before sampling.

\item We develop cross-iteration skip connections at the decoding stage, which promotes efficient propagation of feature information over multiple iterations.

\item Experiments validate that MB-RACS outperforms state-of-the-art approaches, with each component's functionality receiving empirical affirmation. The robustness of measurement-bounds-based rate-adaptive sampling has been effectively demonstrated as well.
\end{itemize}

%% file: Section/related_work.tex
\section{Related Works}\label{sec:related_work}
To our knowledge, current CS methods can be divided into optimization-based methods and data-driven methods. In this section, we will introduce them respectively, and provide a detailed discussion on the works related to our method.

\subsection{Optimization-based Methods}
Traditional CS, which follows an optimization-based method, usually defines the problem of reconstructing the original signal $\textbf{\textit{x}}$ from the known sampled signal $\textbf{\textit{y}}$ as an optimization problem:
\begin{align}
    \hat{\textbf{\textit{x}}} = \argmin_{\textbf{\textit{x}}} \|\boldsymbol{\Phi} \textbf{\textit{x}}-\textbf{\textit{y}}\|_2^2 + \lambda \mathcal{R}(\textbf{\textit{x}}),
\end{align}
where $\mathcal{R}$ denotes a regularization term that capture task-dependent prior knowledge, while $\lambda$ represents the weight of this regularization.

The common methods to solve this optimization problem include: convex optimization methods \cite{doi:10.1137/S003614450037906X}, greedy algorithms \cite{258082}, and gradient descent methods \cite{f957882f5b1e4f3ba0450468093903f1}. 

For image CS, some studies have proposed incorporating image priors into the regularization term. \cite{li2013efficient} introduces total variation to represent local smoothness priors, replacing sparsity priors. \cite{5414429} advocates for sparsity and smoothness in the domain of directional transforms for images. \cite{6341094} presents a collaborative sparsity method, which simultaneously imposes local 2-D and nonlocal 3-D sparsity in an adaptable hybrid domain. \cite{6814320} suggests a group-based sparse representation model, considering inter-patch relationships and adaptively inducing sparsity in natural images within group domains. \cite{7923735} proposes an adaptive bandwise sparsity regularization with patch-specific adaptive PCA for quantized CS.

\subsection{Data-driven Methods}
In recent years, many studies have focused on integrating data-driven methods with CS, achieving commendable reconstruction quality and speed. \cite{Kulkarni_2016_CVPR} proposed the use of CNN to learn the reconstruction from sampled signals to original signals. \cite{8765626} suggested an end-to-end sampling-reconstruction structure, employing CNN to eliminate block effects caused by block sampling. \cite{Zhang_2018_CVPR} replaced the sparsity constraint in the linear transform domain in the traditional optimization-based unfolding algorithm, Iterative Shrinkage-Thresholding Algorithm (ISTA) \cite{doi:10.1137/080716542}, with the sparsity constraint in the network-based nonlinear transform domain. Combining data-driven ideas with traditional unfolding methods has further enhanced model performance. \cite{9019857} continued this approach and explored inter-block relationships. \cite{9298950} extended another unfolding algorithm, AMP, into a network structure. \cite{9934025} proposed introducing Transformers \cite{NIPS2017_3f5ee243} into the reconstruction network to enhance its capacity to model long-distance relationships.
     
Some studies consider the design of deep CS models from other perspectives: the scalability of the model (including the scalability for sampling rates and measurement matrices) and the adaptive allocation of sampling rates across different image blocks. \cite{Shi_2019_CVPR} aims to design a model that adapts to different sampling rates with a single model. It achieves scalability from coarse to fine-grained sampling rates by employing a hierarchical network structure and a greedy algorithm for selecting measurement bases. \cite{9467810} introduced a controllable arbitrary-sampling network method named COAST, which solves the CS problem of any sampling matrix with a single model through a random projection augmentation strategy. \cite{5585813} proposed using the Pulsed Cosine Transform (PCT) to mimic neuronal pulses in the human brain to generate saliency maps to guide the adaptive allocation of sampling rates. Although this method can distinguish which areas are more important to human perception, it cannot rigorously specify and prove what kind of mapping relationship should exist between the importance of an area and the proportion of sampling rate. \cite{9854112} proposed a more efficient network-based saliency map generator for learning adaptive sampling rate allocation strategy. The task of allocating measurements, which is originally a high-dimensional discrete optimization problem, is transformed into a continuous problem and the parameters of the network used for allocating measurements are updated via backpropagation. However, this approach may only find a local optimum rather than the global optimum of the original high-dimensional discrete problem, and have limitations in robustness and generalizability.
   

%% file: Section/proposed_method.tex
\section{Proposed Method}\label{sec:proposed_method}
In this section, we will separately introduce the specific ideas and implementation of the proposed single-stage measurement-bounds-based rate-adaptive sampling, multi-stage rate-adaptive sampling strategy, and decoder-side cross-iteration skip connections.

\input{Section/proposed_method_subsection/measurement-bounds-based_rate-adaptive_sampling}

\input{Section/proposed_method_subsection/multi-stage_rate-adaptive_sampling_strategy}
\input{Section/proposed_method_subsection/cross-iteration_skip_connections}

%% file: Section/proposed_method_subsection/measurement-bounds-based_rate-adaptive_sampling.tex
\begin{figure}[t]
\centering
\begin{minipage}[b]{0.437\dimexpr 0.95\linewidth}
\includegraphics[width=\linewidth, height=\linewidth]{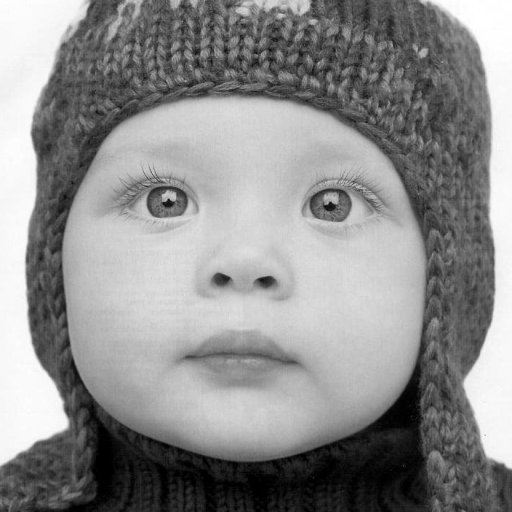} 
\end{minipage}
\hfill
\begin{minipage}[b]{0.563\dimexpr 0.95\linewidth}
\includegraphics[width=\linewidth, height=0.775\dimexpr\linewidth]{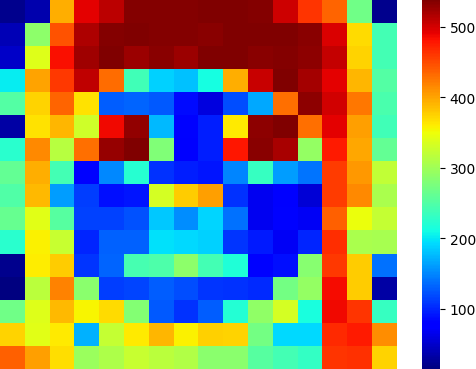} 
\end{minipage}
\caption{The left figure presents an image named ``baby" from the Set5 dataset \cite{BMVC.26.135}, while the right illustrates the distribution of measurements across various blocks, following the measurement bounds proportions and with an overall sampling rate of 0.3. Notably, areas such as the baby's eyes and hat, rich in texture details, are allocated more measurements, whereas smoother regions, like the cheeks, receive fewer due to less detail.}
\label{fig:example_of_measurement_bounds_based_rate-adaptive_sampling}
\end{figure}

\subsection{Single-Stage Measurement-Bounds-based Rate-Adaptive Sampling}\label{sec:3:single-stage measurement-bounds-based rate-adaptive sampling}
When considering how to adaptively allocate sampling rates, we often split the problem into two sub-questions from a human point of view: how do we define the complexity of a block, and how do we assign sampling rate ratios to a set of blocks with different complexities? The traditional CS theory provides well-supported answers to these two questions. The theory mainly characterizes a signal's complexity by its sparsity, which is the count of non-zero elements, within a specific domain. Moreover, the theory ensures that an accurate reconstruction of the signal is possible under the condition:
\begin{align}
    m \geq Ck\log\left(n/k\right), \label{eq:measurement_bounds}
\end{align}
as referred to in Sec.~\ref{sec:introduction}. Here, $C$ is a constant, $k$ is the sparsity of the signal, $n$ is the length of the signal, and $m$ represents the number of measurements. The term $Ck\log\left(\frac{n}{k}\right)$ is known as the measurement bounds. It is clear that the measurement bounds establish a correlation from the complexity of the signal, $k$, to the quantity of measurements allocated to it. As a result, the previously proposed questions find satisfactory solutions within traditional CS theory. 

\begin{figure*}[t]
\centering
\includegraphics[width=\textwidth]{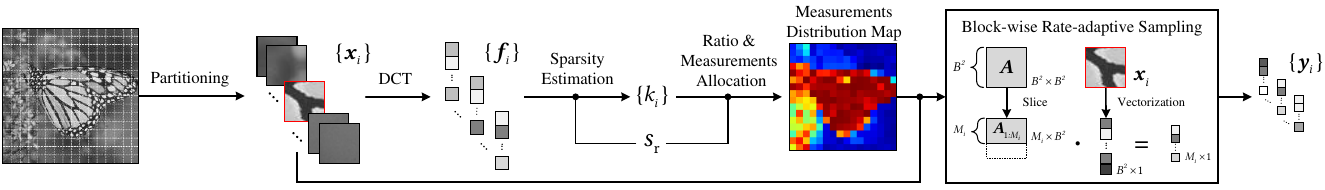}
\caption{Illustration of single-stage measurement-bounds-based rate-adaptive sampling, comprising partitioning, DCT transformation, block-wise sparsity estimation, ratio and measurements allocation, and block-wise rate-adaptive sampling.}
\label{fig:measurement-bounds-based_rate-adaptive_sampling_illustration}
\end{figure*}

Fig.~\ref{fig:example_of_measurement_bounds_based_rate-adaptive_sampling} illustrates an example of measurement-bounds-based sampling rate allocation. As mentioned in the following text, we calculate the measurement bounds using the sparsity of the image block in the Discrete Cosine Transform (DCT) domain. It can be observed that the blocks with more complex textures are allocated a greater number of measurements. In the following, we will introduce the two steps for implementing single-stage measurement-bound-based rate-adaptive sampling, with Fig.~\ref{fig:measurement-bounds-based_rate-adaptive_sampling_illustration} illustrating the associated sampling process.

\begin{figure}[t]
\centering
\includegraphics[width=\linewidth]{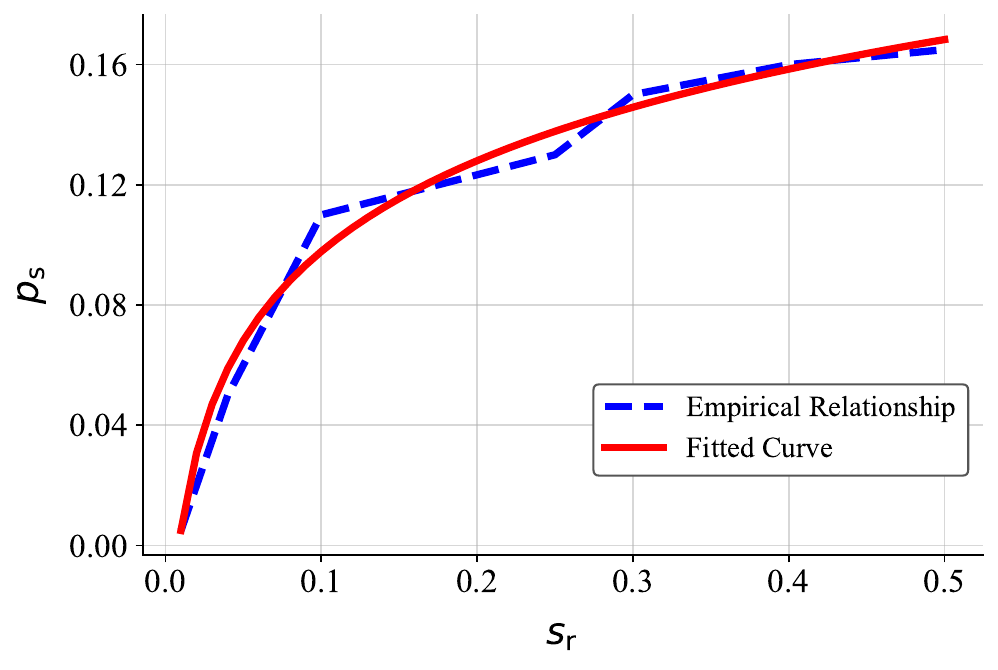}
\caption{Empirical and logarithmically fitted trends for optimal $(s_\text{r}, p_\text{s})$ pairs in model performance.}
\label{fig:s_r_p_s}
\end{figure}

\subsubsection{Determining Sparsity Threshold Based on Overall Sparsity Ratio}
\label{sec:3:single-stage measurement-bounds-based rate-adaptive sampling:sparsity threshold}
Initially, we consider an image $\textbf{\textit{x}} \in \mathbb{R}^{H\times W}$, with $H$ and $W$ signifying the image's width and height respectively. The image is partitioned into discrete, non-overlapping blocks of size $B\times B$. We assume that $H$ and $W$ are divisible by $B$. If not, we can achieve divisibility of the image's dimensions by the block's side length by appending zeros along the image's borders. We denote these blocks as $\textbf{\textit{x}}_i \in \mathbb{R}^{B \times B}, \ i = 1, 2, \ldots, \frac{HW}{B^2}$. Subsequently, a DCT transformation $\text{dct}()$ is applied to the $i^{\text{th}}$ block, yielding its frequency coefficient $\textbf{\textit{f}}_i$:
\begin{align}
\textbf{\textit{f}}_i=\text{DCT}(\textbf{\textit{x}}_i).
\end{align}
After the DCT transformation, a multitude of coefficients tend towards zero without being exactly zero. Therefore, setting a sparsity threshold $T$ becomes essential, leading to coefficients with absolute values below this threshold being assigned a zero value. Next, we obtain the relationship between the sparsity threshold and the overall sparsity ratio $p_\text{s}$:
\begin{align}
    p_\text{s} = \frac{\sum_{i}\|\max(\abs{\textbf{\textit{f}}_i}-T, 0)\|_0}{HW}, \label{eq:overall_sparsity_ratio}
\end{align}
where $\|\|_0$ symbolizes the zero-norm, indicating the quantity of non-zero values in the vector, and $p_\text{s}$ represents the proportion of non-zero elements.

According to \eqref{eq:measurement_bounds}, we believe that an increase in sparsity will lead to an increase in the required number of measurements if we still want to achieve a good reconstruction. From another perspective, when we change the total number of measurements, we may also need to alter the calculation method for the sparsity of each block by changing the sparsity threshold, so that the sum of the measurement bounds for all blocks matches the given total number of measurements. Intuitively, when the overall sampling rate $s_\text{r}$ increases, the overall sparsity ratio should also increase. Next, we explore the potential correspondence between these two quantities through experiments. This is visually represented in the ``Empirical Relationship" in Fig.~\ref{fig:s_r_p_s}, which shows the $p_\text{s}$ values that yield better model performance at different given overall sampling rates $s_\text{r}$. The progression of the curve confirms our intuition, indicating that a more substantial $s_\text{r}$ value corresponds to a larger $p_\text{s}$ value, and the form of the curve closely resembles a logarithmic function. Thus, these $(s_\text{r}, p_\text{s})$ pairs are fit using the function $p_\text{s}=b\log(a(s_\text{r}-s_{\text{r};1})+1)+p_{\text{s};1}$, where $a,b$ are the parameters to be fitted, and $(s_{r;1}, p_{s;1})$ symbolizes the pair with the least $s_\text{r}$ value amongst all tested $(s_\text{r}, p_\text{s})$ pairs. The ``Fitted Curve" in Fig.~\ref{fig:s_r_p_s} presents the fitting outcome, illustrating a general alignment in the trends of the fitted and the empirically drawn curves. 

Thus, in setting the sparsity threshold, we initially identify the corresponding $p_\text{s}$ for a specified $s_\text{r}$ based on the fitted curve. Subsequently, we determine the value of $T$ based on \eqref{eq:overall_sparsity_ratio} when $p_\text{s}$ is given. Specifically for this step of determining $T$, given the limited search space and the lack of high-precision solution necessity, we employ an exhaustive search for the resolution.

\subsubsection{Distribution of Measurements}
After determining the sparsity threshold $T$, we can use it to estimate the sparsity $k_i$ of the $i^{\text{th}}$ block:
\begin{align}
    k_i=\|\max(\abs{\textbf{\textit{f}}_i}-T, 0)\|_0.
\end{align}
Subsequently, using \eqref{eq:measurement_bounds}, we can determine the measurement bounds for the $i^{\text{th}}$ block while disregarding its constant term:
\begin{align}
    m_i = k_i\log_{10}(B^2/k_i). \label{eq:estimated_measurement_bounds}
\end{align}
Following that, we can distribute the sampling rate $s_i$ and the number of measurements $M_i$ for the $i^{\text{th}}$ block based on the previously determined measurement bounds:
\begin{align}
    s_i&=\eta\frac{m_i}{\sum_{i}m_i}s_\text{r}\frac{HW}{B^2}, \label{eq:distribute sampling rate}\\
    M_i&=\round{s_iB^2}.
\end{align}
In the computation of $s_i$, the measurement bounds occur in both the numerator and denominator, leading to the cancellation of constant term in \eqref{eq:measurement_bounds}, justifying the omission of this term in \eqref{eq:estimated_measurement_bounds}.
As there might be errors between the actual sampling rate, $\frac{\sum_iM_i}{HW}$, and the expected sampling rate $s_\text{r}$ due to rounding, we introduce the variable $\eta$ to slightly adjust $s_i$, ensuring the actual sampling rate aligns with the expected one.

After determining the number of measurements for each block, we initialize a learnable measurement matrix $\textbf{\textit{A}} \in \mathbb{R}^{B^2\times B^2}$ to measure all the blocks:
\begin{align}
\textbf{\textit{y}}_i=\textbf{\textit{A}}_{1:M_i}\mathcal{V}(\textbf{\textit{x}}_i).
\end{align}
In this equation, $\textbf{\textit{A}}_{p:q}$ refers to extracting the $p^{\text{th}}$ through $q^{\text{th}}$ rows from $\textbf{\textit{A}}$ to serve as the measurement matrix.
$\mathcal{V}()$ signifies the process of vectorizing a two-dimensional block into a one-dimensional vector, while $y_i$ is the sampled signal for the $i^{\text{th}}$ block.

%% file: Section/proposed_method_subsection/multi-stage_rate-adaptive_sampling_strategy.tex
\subsection{Multi-Stage Rate-Adaptive Sampling Strategy}\label{sec:3:multi-stage rate-adaptive sampling strategy}

\begin{figure*}[t]
\centering
\includegraphics[width=\textwidth]{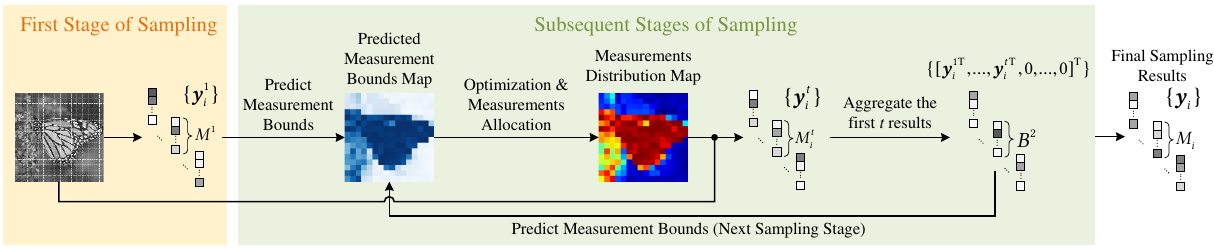}
\caption{Illustration of the multi-stage rate-adaptive sampling strategy: the first sampling stage employs rate-fixed sampling, with subsequent sampling stages involving measurement bounds prediction, optimization of allocation ratio, measurements allocation, and rate-adaptive sampling.}
\label{fig:multi-stage_rate-adaptive_sampling_strategy}
\end{figure*}

The aforementioned method assumes that we can access sparsity information of the signal prior before sampling. However, this is not always feasible in certain practical applications. Taking single-pixel imaging \cite{4472247} as an example, if we consider photography as a sampling process, we cannot obtain information about the original image through other methods before sampling. To overcome this, we introduce a multi-stage rate-adaptive sampling strategy, partitioning the sampling process into several stages. Leveraging information from previous samplings, we estimate the measurement bounds for each block. This estimation then informs the allocation of the sampling rate and the count of measurements for the succeeding sampling stage. Fig.~\ref{fig:multi-stage_rate-adaptive_sampling_strategy} illustrates the process of the multi-stage rate-adaptive sampling strategy.

\subsubsection{Sampling Process}
To elaborate, assume that we partition the sampling process into $N$ stages, where each stage has an approximate sampling rate of $s_\text{r}/N$, allowing for rounding deviations.
In the first stage, each block is sampled at an identical rate, given that we haven't yet gathered any information about the image. For the remaining $N-1$ stages, we predict the measurement bounds based on the previous sampled signals.

\paragraph{Sampling in the First Stage}
We set the sampling rate for the first stage $s_\text{r}^1=s_\text{r}/N$. For the $i^{\text{th}}$ block, the count of measurements in this stage is given by $M_i^1=M^1=\floor{s_\text{r}^1B^2}$. We then proceed to sample the $i^{\text{th}}$ block using the initial $M^1$ rows of $\textbf{\textit{A}}$:
\begin{align}
    \textbf{\textit{y}}_i^1 = \textbf{\textit{A}}_{1:M^1}\mathcal{V}(\textbf{\textit{x}}_i).
\end{align}

\paragraph{Sampling in Subsequent Stages}
\label{sec:3:multi-stage rate-adaptive sampling strategy:sampling process:sampling in subsequent stages}
Here, we are concerned with the sampling rate and measurements allocation for the $t^{\text{th}}$ (where $t > 1$) sampling stage. The actual 
cumulative sampling rate for the first $t-1$ stages is given by $\sum_i\sum_{j=1}^{t-1}M_i^j/HW$, where $M_i^j$ represents the actual number of measurements allocated to the $i^{\text{th}}$ block during the $j^{\text{th}}$ stage. Moreover, the expected 
cumulative sampling rate up to the $t^{\text{th}}$ sampling stage is set to $ts_\text{r}/N$. Thus, the expected sampling rate specifically for the $t^{\text{th}}$ stage $s_\text{r}^t=ts_\text{r}/N-\sum_i\sum_{j=1}^{t-1}M_i^j/HW$. Notably, by this stage, we already have the sampling results of the initial $t-1$ stages for each block. Therefore, we designed a measurement bounds prediction network that uses these sampling results to predict the measurement bounds of these blocks:
\begin{align}
g_{\boldsymbol{\theta}_m}\left(\left[{\textbf{\textit{y}}_i^1}^\transpose, \ldots, {\textbf{\textit{y}}_i^{t-1}}^\transpose, 0, \ldots, 0\right]^\transpose\right)=\hat{m}_i^t.
\end{align}
The measurement bounds prediction network $g_{\boldsymbol{\theta}_m}$ consists of several CNN layers and residual blocks \cite{He_2016_CVPR}, where $\boldsymbol{\theta}_m$ represents the network parameters.  $\textbf{\textit{y}}_i^1, \ldots, \textbf{\textit{y}}_i^{t-1}$ represent the sampling signals obtained from the first $t-1$ sampling stages for the $i^{\text{th}}$ block. Since the length of the concatenated sampling signal is likely to differ for each block, we pad zeros to the end of each concatenated sampling signal to ensure a consistent length of $B^2$. We compute the Kullback-Leibler (KL) Divergence-based loss to measure the discrepancy when the predicted measurement bounds $\hat{m}_i^t$ are employed to approximate the actual measurement bounds $m_i$, both in terms of distribution ratios, and use this KL divergence-based loss for network training:
\begin{align}
    \mathcal{L}_{\text{KL}}=\frac{1}{(N-1)\frac{HW}{B^2}}\sum_{t=2}^N\sum_i-\frac{m_i}{\sum_im_i}\log\frac{\hat{m}_i^t}{\sum_i\hat{m}_i^t}, 
    \label{eq:KL loss for measurement bounds prediction network training}
\end{align}
where the normalization factor of $\frac{1}{(N-1)\frac{HW}{B^2}}$ is chosen to average the loss over all sampling stages that have the step of predicting measurement bounds and over all blocks.

Upon we obtain the predicted measurement bounds, we set the expected proportion of measurements allocated to the $i^{\text{th}}$ block after the $t^{\text{th}}$ stage as $p_i^t = \frac{\hat{m}_i^t}{\sum_i \hat{m}_i^t}$. However, the expected proportion is likely not achievable due to the fixed allocation ratios in the first $t-1$ stages. Therefore, we can only adjust the allocation ratio for the current $t^{\text{th}}$ stage to make the actual cumulative allocation ratio over the first $t$ stages approximate the expected one. Sec. \ref{sec:3:multi-stage rate-adaptive sampling strategy:subproblem of distribution ratio optimization} describes the specific distribution ratio optimization method for allocating the sampling rate for the current $t^{\text{th}}$ stage. We define the allocation ratio of the $i^{\text{th}}$ block, optimized using the aforementioned method, as ${q_i^t}^{\star}$.

Once we determine the optimized allocation ratio, we can allocate the sampling rate and the number of measurements to the $i^\text{th}$ block based on this allocation ratio:
\begin{align}
    s_i^t & = \eta {q_i^t}^{\star}s_\text{r}^t\frac{HW}{B^2},\\
    M_i^t & = \round{s_i^tB^2}.
\end{align}
Here, $\eta$ plays a similar role as in \eqref{eq:distribute sampling rate}, its primary function being to align the actual sampling rate closely with the desired one. Next, we extract rows $\sum_{j=1}^{t-1}M_i^j+1$ through $\sum_{j=1}^{t}M_i^j$ from measurement matrix $\textit{\textbf{A}}$ to sample the $i^\text{th}$ block:
\begin{align}
    \textit{\textbf{y}}_i^t = \textit{\textbf{A}}_{\sum_{j=1}^{t-1}M_i^j+1:\sum_{j=1}^{t}M_i^j}\mathcal{V}(\textit{\textit{x}}_i).
\end{align}
After completing $N$ stages of sampling, we obtained the total number of measurements  as well as the final sampling result for the $i^{\text{th}}$ block:
\begin{align}
    M_i & = \sum_{j=1}^{N}M_i^j, \\
    \textit{\textbf{y}}_i & = \left[{\textit{\textbf{y}}_i^1}^\transpose, \ldots, {\textbf{\textit{y}}_i^{N}}^\transpose\right]^\transpose.
\end{align}

\subsubsection{Subproblem of Distribution Ratio Optimization}
\label{sec:3:multi-stage rate-adaptive sampling strategy:subproblem of distribution ratio optimization}

\begin{figure}[t]
\centering
\includegraphics[width=\linewidth]{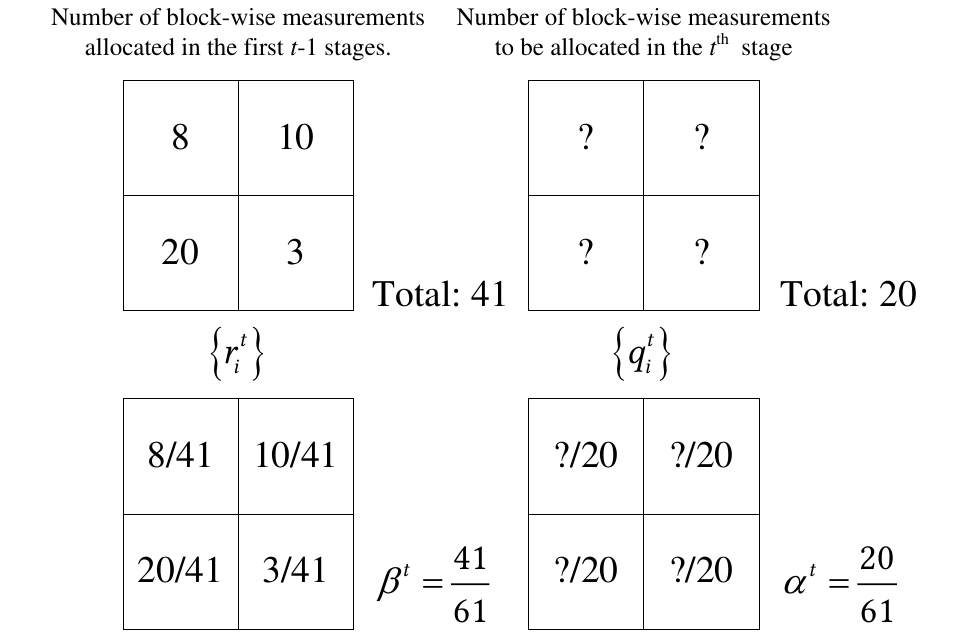}
\caption{A toy example of $r_i^t$, $q_i^t$, $\beta^t$, and $\alpha^t$.}
\label{fig:distribution ratio optimization toy example}
\end{figure}

\paragraph{Problem Description}
In the following, we will define some symbols and describe the distribution ratio optimization problem as a convex optimization problem. The proportion of measurements yet to be allocated in the current $t^{\text{th}}$ stage and the proportion of measurements allocated in the first $t-1$ stages, with respect to the total number of allocatable measurements in the first $t$ stages, are given by $\alpha^t = \frac{s_\text{r}^t}{ts_\text{r}/N}$ and $ \beta^t = 1 - \alpha^t = \frac{\sum_i\sum_{j=1}^{t-1}M_i^j/HW}{ts_\text{r}/N}$, respectively. The fixed allocation ratio for the $i^{\text{th}}$ block is $r_i^t=\frac{\sum_{j=1}^{t-1}M_i^j}{\sum_i\sum_{j=1}^{t-1}M_i^j}$. Suppose the allocation ratio to be determined in the $t^{\text{th}}$ stage is $\{q_i^t\}$. Therefore, after the $t^{\text{th}}$ sampling stage, the actual allocation ratio is $\{\alpha^t q_i^t+\beta^t r_i^t\}$. Fig.~\ref{fig:distribution ratio optimization toy example} provides a toy example depicting $r_i^t$, $q_i^t$, $\beta^t$, and $\alpha^t$. Our objective is to adjust $\{q_i^t\}$ so that $\{\alpha^t q_i^t+\beta^t r_i^t\}$ is close to $\{p_i^t\}$ under the measurement of KL divergence-based object function. Additionally, $q_i^t$ is subject to upper and lower bounds. Clearly, the lower bound of $q_i^t$ is 0. Moreover, given that the number of measurements for the $i^{\text{th}}$ block must not exceed the length of its original signal, and considering that the count of measurements for the $i^{\text{th}}$ block at the $t^{\text{th}}$ sampling stage is approximately $q_i^ts_\text{r}^tHW$ (accounting for a rounding error), we can derive the following relationship:
\begin{align}
    q_i^ts_\text{r}^tHW + \sum_{j=1}^{t-1}M_i^j \leq B^2.
\end{align}
Therefore, we can obtain the upper bound of $q_i^t$, $a_i^t=\frac{B^2-\sum_{j=1}^{t-1}M_i^j}{s_\text{r}^tHW}$. Finally, we integrate the above considerations into a convex optimization problem:
\begin{equation}
\begin{aligned}
\argmin_{\{q_i^t\}} \quad  - & \sum_ip_i^t\log (\alpha^t q_i^t+\beta^t r_i^t) \\
\text{subject to} \quad & \sum_i q_i^t = 1, \\
                        & 0 \leq q_i^t \leq a_i^t \quad \text{for all} \quad i. 
\label{eq:allocation ratio optimization problem}                 
\end{aligned}
\end{equation}

\paragraph{Obtaining the Implicit Solution of the Optimization Problem Using the Lagrangian Multiplier  Method}

We employ the Karush-Kuhn-Tucker (KKT) conditions based on the Lagrange multiplier method to resolve this issue, yielding an implicit solution. The detailed procedure is provided in Appendix~A.

\paragraph{Solving for the Lagrange Multiplier to Obtain Explicit Solution}
We first frame the solution of the Lagrange multiplier as the problem of finding the root of a function. Building on our earlier discussion, we have derived the closed-form solution of the convex optimization problem as: ${q_i^t}^{\star} = \min\left(\max\left(\frac{p_i^t}{\nu^{\star}} - \frac{\beta^t r_i^t}{\alpha^t}, 0\right), a_i^t\right)$. The next step is to identify the value of the Lagrange multiplier $\nu^{\star}$ in order to determine the solution for ${q_i^t}^{\star}$. To simplify our notation, we substitute $\nu^\star$ with $\frac{1}{\mu^\star}$ (based on the conclusion in Appendix~A that $\nu^\star > 0$, we only consider cases where $\mu^\star > 0$). This leads us to rewrite the closed-form solution of the convex optimization problem as:
\begin{align}
{q_i^t}^{\star} = \min\left(\max\left(\mu^{\star}p_i^t - \frac{\beta^t r_i^t}{\alpha^t}, 0\right), a_i^t\right).
\label{eq:distribution ratio optimization}
\end{align}
Define $Q_i^t(\mu) = \min\left(\max\left(\mu p_i^t- \frac{\beta^t r_i^t}{\alpha^t}, 0\right), a_i^t\right)$, and let $Q^t(\mu) = \sum_i Q_i^t(\mu)$. Based on the constraint condition in \eqref{eq:allocation ratio optimization problem} that the sum of the allocation ratios should be equal to 1, our problem is transformed into finding $\mu$ that satisfies $Q^t(\mu) = 1$.

\begin{figure}
    \centering
    \includegraphics[width=0.48\textwidth]{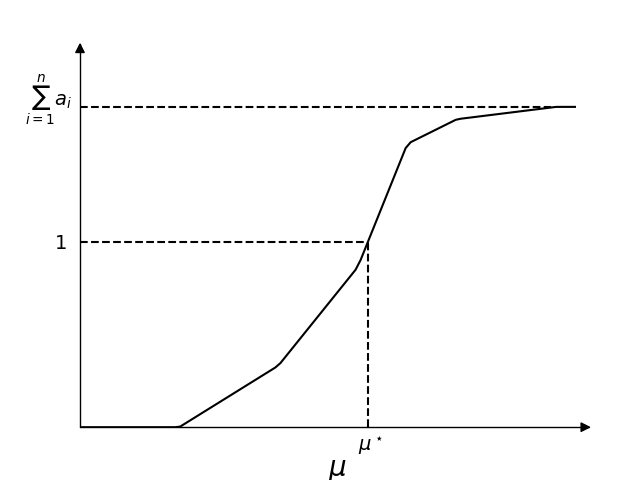}
    \caption{The approximate graphical representation of the function $Q^t(\mu)$.}
    \label{fig:the approximate shape of Q(mu)}
\end{figure}

Before examining the function $Q^t(\mu)$, let us first define some notation: $N_\text{l}^t(\mu)=\{i|Q_i^t(\mu)=0\}$, $N_\text{c}^t(\mu)=\{i|0<Q_i^t(\mu)<a_i^t\}$ and $N_\text{u}^t(\mu)=\{i|Q_i^t(\mu)=a_i^t\}$. With the help of this notation, we obtain another representation of $Q^t(\mu)$:
\begin{align}
    Q^t(\mu) = \sum_i Q_i^t(\mu) &= \sum_{i\in N_\text{c}^t(\mu)}(\mu p_i^t - \frac{\beta^t r_i^t}{\alpha^t}) + \sum_{i\in N_\text{u}^t(\mu)}a_i^t.
\end{align}
The derivative of $Q^t(\mu)$ with respect to $\mu$ can be formulated as:
\begin{align}
    {Q^t}'(\mu) &= \sum_{i\in N_\text{c}^t(\mu)}p_i^t.
\end{align}
The function $Q^t(\mu)$ should be a piecewise linear monotonically increasing function with its domain being $(0, +\infty)$ and its range being $[0, \sum_{i=1}^n a_i^t]$. The boundary points of each linear segment correspond to those points $\mu$ that cause changes in $\{N_\text{J}(\mu) | \text{J} = \text{l}, \text{c}, \text{u}\}$. Fig.~\ref{fig:the approximate shape of Q(mu)} shows the approximate shape of the $Q^t(\mu)$ function.

Next, we combine Newton's method with the bisection method to solve for the root of $Q^t(\mu) = 1$, ensuring efficient computation while guaranteeing convergence. Initially, we solve iteratively using Newton's method. If convergence is not achieved, we use the bisection method to ensure that the iterative solution remains within the known upper and lower bounds of the root.

\input{Section/proposed_method_subsection/newton_and_bisection_algorithm}

Specifically, the principle behind Newton's method is that during each iteration, it approximates the function as being locally linear around the current estimate and subsequently updates the root approximation based on this linear assumption. In the $\tau^{\text{th}}$ iteration, we have:
\begin{align}
    1&=(\mu^{(\tau+1)}-\mu^{(\tau)}){Q^t}'(\mu^{(\tau)})+Q^t(\mu^{(\tau)}),
\end{align}
where $\mu^{(\tau)}$ is the result of the $\tau^{\text{th}}$ iteration of Newton's method. By relocating $\mu^{(\tau+1)}$ to the left side of the equation and shifting the other terms to the right side, we obtain:
\begin{align}
    & \mu^{(\tau+1)} = \frac{1-Q^t(\mu^{(\tau)})}{{Q^t}'(\mu^{(\tau)})}+\mu^{(\tau)} \nonumber \\
    & =\frac{1-\sum_{i\in N_\text{c}^t(\mu^{(\tau)})}(\mu^{(\tau)}p_i^t-\frac{\beta^t r_i^t}{\alpha^t})-\sum_{i\in N_\text{u}^t(\mu^{(\tau)})}a_i^t}{\sum_{i\in N_\text{c}^t(\mu^{(\tau)})}p_i^t}+\mu^{(\tau)} \nonumber \\
    & =\frac{1+\sum_{i\in N_\text{c}^t(\mu^{(\tau)})}\frac{\beta^t r_i^t}{\alpha^t}-\sum_{i\in N_\text{u}^t(\mu^{(\tau)})}a_i^t}{\sum_{i\in N_\text{c}^t(\mu^{(\tau)})}p_i^t}. \label{eq:newton_iter}
\end{align}
In general, both the numerator and the denominator in the last expression of \eqref{eq:newton_iter} are greater than zero, unless the initial value for Newton's method is set too small or too large. Empirically, setting $\mu^{(0)} = \frac{\beta^t}{\alpha^t}$ ensures both are positive. We set the termination condition as $\{N_\text{J}^t(\mu^{(\tau)}) | \text{J} = \text{l}, \text{c}, \text{u}\}$ being entirely equal to $\{N_\text{J}^t(\mu^{(\tau+1)}) | \text{J} = \text{l}, \text{c}, \text{u}\}$, that is, $\mu^{(\tau)}$ and $\mu^{(\tau+1)}$ are in the same linear function segment. In this case, we can deduce that $\mu^{(\tau+1)}$ is the root of $Q^t(\mu) = 1$:
\begin{equation}
\begin{aligned}
    Q^t(\mu^{(\tau+1)}) & = \sum_{i\in N_\text{c}^t(\mu^{(\tau+1)})}(\mu^{(\tau+1)}p_i^t - \frac{\beta^t r_i^t}{\alpha^t}) + \sum_{i\in N_\text{u}^t(\mu^{(\tau+1)})}a_i^t \\
    & = \sum_{i\in N_\text{c}^t(\mu^{(\tau)})}(\mu^{(\tau+1)}p_i^t - \frac{\beta^t r_i^t}{\alpha^t}) + \sum_{i\in N_\text{u}^t(\mu^{(\tau)})}a_i^t \\
    & \stackrel{\text{According to \eqref{eq:newton_iter}}}{=}1. \nonumber
\end{aligned}
\end{equation}

\begin{figure}[t]
\centering
\includegraphics[width=\linewidth]{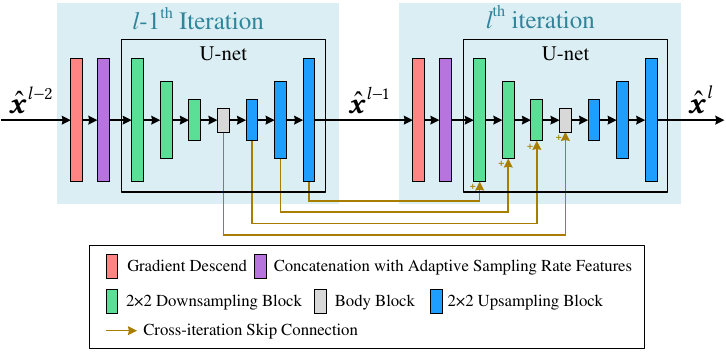}
\caption{Illustration of the cross-iteration skip connections: we add the features output from the three upsampling blocks and the body block of the U-net in ${l-1}^\text{th}$ iteration to the features in the three downsampling blocks and the body block in the U-net of ${l}^\text{th}$ iteration.}
\label{fig:cross-iteration_skip_connection}
\end{figure}

Subsequently, we wish to consider the convergence issue of Newton's method under discussion. The difficulty in directly referencing previous methods of proving the convergence of Newton's method is that the general Newton's method deals with smooth functions, and considers the convergence speed of the error upper bound as $\tau$ increases. However, in this specific scenario, during the Newton's iteration process, $\mu^{(\tau)}$ will fall into different linear segments, until $\mu^{(\tau)}$ falls into the linear segment identical to the root (as stated above, once we find the linear segment where the root is located, we can obtain the root through one iteration of Newton's method). From this perspective, this problem is concerned with finding the root's linear segment from all possible linear segments, rather than focusing on the convergence speed of the error in the solution. 

Consequently, we are currently seeking to propose a fallback plan for when Newton's method fails to converge, ensuring that we eventually identify the linear segment hosting the root. We first present Lemma~B.1 in Appendix~B. This lemma allows us to understand the interval in which $\mu^\star$ is located and assists us further narrow down this interval. Specifically, we can determine the upper and lower bounds of $\mu^\star$ by comparing the relationship between $\mu^{(\tau)}$ and $\mu^{(\tau+1)}$. For example, if $\mu^{(\tau)} < \mu^{(\tau+1)}$, based on Lemma~B.1, we can deduce that $\mu^{(\tau)} < \mu^{\star}$. Then, compare $\mu^{(\tau)}$ with the existing lower bound; if $\mu^{(\tau)}$ is greater than the current lower bound, update the lower bound to $\mu^{(\tau)}$.

\input{Section/experiments_and_results_section/tables/single-stage_table_5_datasets_different_methods_results}
\input{Section/experiments_and_results_section/tables/single-stage_table_2_datasets_different_methods_results}

Next, we can assess the convergence at each iteration of Newton's method by using the upper and lower bounds obtained from Lemma~B.1. 
If the result of a particular iteration surpasses the upper or lower bound, we consider that this iteration has not converged. Therefore, we utilize the bisection method to replace the result of this iteration by the average of the lower and upper bounds. Even if all iterations in Newton's method fail to converge, we can still use Lemma~B.1 to ascertain whether the root is smaller or larger than our current result. This helps define the bisection method's search interval, guiding the overall process to locate the linear segment where the root lies via bisection method instead of Newton's method. We have developed Algorithm \ref{alg:find root with newton's method and bisection method} based on this methodology. For the sake of convenient notation, we have removed the superscript $t$ that indicates the stage number. Additionally, 
\begin{align}
\chi_\textbf{\textit{b}}(\bm{\omega})_i= \begin{cases} 1 & \text{if } \omega_i=b_i \\ 0  & \text{if } \omega_i\neq b_i
\end{cases},
\end{align}
where $\chi_\textbf{\textit{b}}(\bm{\omega})_i$ is the ${i}^\text{th}$ component of $\chi_\textbf{\textit{b}}(\bm{\omega})$, which maps from $:\mathbb{R}^{\frac{HW}{B^2}}$ to $\mathbb{R}^{\frac{HW}{B^2}}$($\textbf{\textit{b}}\in \mathbb{R}^{\frac{HW}{B^2}}$).

%% file: Section/proposed_method_subsection/newton_and_bisection_algorithm.tex
\begin{algorithm}[t]
\caption{Find $\mu^{\star}$ and $\textit{\textbf{q}}^{\star}$ using Newton's Method and Bisection Method}
\label{alg:find root with newton's method and bisection method}
\begin{algorithmic}[1]
\INPUT {$\textit{\textbf{p}}, \textit{\textbf{r}}, \textit{\textbf{a}}, \alpha, \beta$}
\OUTPUT {$\mu^{\star}, \textit{\textbf{q}}^{\star}$}
\State $B_\text{l} \gets \frac{\beta}{\alpha}$ 
\State $B_\text{u} \gets \text{MAXSIZE}$
\State $\mu_{\text{old}} \gets \frac{\beta}{\alpha}$
\State $\textbf{\textit{q}} \gets \min\left(\max\left(\mu_{\text{old}}\cdot \textbf{\textit{p}}-\frac{\beta\textit{\textbf{r}}}{\alpha}, \textbf{0}\right), \textbf{\textit{a}}\right)$
\State $\textbf{\textit{m}}_\text{u} \gets \chi_{\textbf{\textit{a}}}(\textbf{\textit{q}})$
\State $\textbf{\textit{m}}_\text{l} \gets \chi_\textbf{0}(\textbf{\textit{q}})$
\While{True}
    \State $\textbf{\textit{m}}_\text{c} \gets (\textbf{1}-\textbf{\textit{m}}_\text{l})\odot(\textbf{1}-\textbf{\textit{m}}_\text{u})$
    \State $\mu \gets \frac{1+\text{SUM}(\textbf{\textit{m}}_\text{c}\odot\textit{\textbf{r}})\beta/\alpha-\text{SUM}(\textbf{\textit{m}}_\text{u}\odot\textbf{\textit{a}})}{\text{SUM}(\textbf{\textit{p}}\odot \textbf{\textit{m}}_\text{c})}$
    \If{$\mu_{\text{old}} < \mu$}
        \If{$B_\text{l} < \mu_{\text{old}}$}
            \State $B_\text{l} \gets \mu_{\text{old}}$
        \EndIf
    \ElsIf{$\mu_{\text{old}} > \mu$}
        \If{$B_\text{u} > \mu_{\text{old}}$}
            \State $B_\text{u} \gets \mu_{\text{old}}$
        \EndIf
    \EndIf
    \If{$\mu<=B_\text{l}$ or $\mu>=B_\text{u}$}
        \State $\mu \gets (B_\text{l}+B_\text{u})/2$
    \EndIf
    \State $\mu_{\text{old}} \gets \mu$
    \State $\textbf{\textit{q}} \gets \min\left(\max\left(\mu\cdot \textbf{\textit{p}}-\frac{\beta\textit{\textbf{r}}}{\alpha}, \textbf{0}\right), \textbf{\textit{a}}\right)$
    \State $\textbf{\textit{m}}_\text{u}^{'} \gets \chi_{\textbf{\textit{a}}}(\textbf{\textit{q}})$
    \State $\textbf{\textit{m}}_\text{l}^{'} \gets \chi_\textbf{0}(\textbf{\textit{q}})$
    \If {$\textbf{\textit{m}}_\text{u}$ $=$ $\textbf{\textit{m}}_\text{u}^{'}$ and $\textbf{\textit{m}}_\text{l}$ $=$ $\textbf{\textit{m}}_\text{l}^{'}$}
        \State \Return $\mu$, $\textbf{\textit{q}}$
    \Else
        \State $\textbf{\textit{m}}_\text{u} \gets \textbf{\textit{m}}_\text{u}^{'}$
        \State $\textbf{\textit{m}}_\text{l} \gets \textbf{\textit{m}}_\text{l}^{'}$
    \EndIf
\EndWhile
\end{algorithmic}
\end{algorithm}

%% file: Section/experiments_and_results_section/tables/single-stage_table_5_datasets_different_methods_results.tex
\begin{table*}
\centering
\caption{PSNR and SSIM comparisons for MB-RACS and 9 baselines on Set11, CBSD68, Set5, and Set14 datasets at sampling rates of 0.01, 0.04, 0.1, 0.25, 0.3, 0.4, and 0.5. The subscript ``ss'' denotes the use of the single-stage rate-adaptive sampling method. The subscript ``ms;$x$'' denotes the use of the $x$-stage rate-adaptive sampling method.}
\label{tab:comparison with state-of-the-arts methods on five datasets}
\resizebox{\linewidth}{!}{
\begin{tabular}{c c c c c c c c c c c c c c c c}
\toprule
\multirow{2}{*}[-1em]{Dataset} & \multirow{2}{*}[-1em]{Method} & \multicolumn{14}{c}{Sampling Ratio} \\ \cmidrule(lr){3-16} 
& & \multicolumn{2}{c}{0.01} & \multicolumn{2}{c}{0.04} & \multicolumn{2}{c}{0.1} & \multicolumn{2}{c}{0.25} & \multicolumn{2}{c}{0.3} & \multicolumn{2}{c}{0.4} & \multicolumn{2}{c}{0.5} \\
\cmidrule(lr){3-4} \cmidrule(lr){5-6} \cmidrule(lr){7-8} \cmidrule(lr){9-10} \cmidrule(lr){11-12} \cmidrule(lr){13-14} \cmidrule(lr){15-16} & & PSNR & SSIM & PSNR & SSIM & PSNR & SSIM & PSNR & SSIM & PSNR & SSIM & PSNR & SSIM & PSNR & SSIM \\
\midrule
\specialrule{0em}{4pt}{1pt}
\midrule
\multirow{11}{*}[-3.2em]{Set11} & CSNet & 20.91 & 0.5504 & 24.89 & 0.7492 & 28.31 & 0.8590 & 33.07 & 0.9360 & 34.34 & 0.9487 & 36.41 & 0.9637 & 37.92 & 0.9723 \\ \cmidrule(l){2-16}
 & ISTA-Net$^+$ & 21.41 & 0.5776 & 25.80 & 0.7802 & 29.57 & 0.8784 & 34.82 & 0.9479 & 36.20 & 0.9588 & 38.46 & 0.9715 & 40.32 & 0.9799 \\ \cmidrule(l){2-16}
 & Hybrid-LISTA & 21.45 & 0.5720 & 25.37 & 0.7607 & 28.84 & 0.8638 & 33.18 & 0.9346 & 34.48 & 0.9483 & 36.88 & 0.9662 & 39.12 & 0.9776 \\ \cmidrule(l){2-16}
 & SCSNet & 21.04 & 0.5616 & 25.12 & 0.7619 & 28.48 & 0.8647 & 33.33 & 0.9383 & 34.60 & 0.9504 & 36.62 & 0.9651 & 38.07 & 0.9731 \\ \cmidrule(l){2-16}
 & OPINE-Net & 21.40 & 0.5795 & 25.92 & 0.7866 & 29.82 & 0.8831 & 35.26 & 0.9518 & 36.54 & 0.9607 & 38.79 & 0.9729 & 40.78 & 0.9808 \\ \cmidrule(l){2-16}
 & AMP-Net & 21.43 & 0.5776 & 25.76 & 0.7774 & 29.56 & 0.8782 & 34.88 & 0.9486 & 36.18 & 0.9584 & 38.50 & 0.9716 & 40.50 & 0.9802 \\ \cmidrule(l){2-16}
 & COAST & 21.33 & 0.5747 & 25.77 & 0.7798 & 29.63 & 0.8789 & 34.86 & 0.9484 & 36.13 & 0.9581 & 38.33 & 0.9711 & 40.23 & 0.9795 \\ \cmidrule(l){2-16}
 & TransCS & 21.29 & 0.5587 & 25.79 & 0.7778 & 29.77 & 0.8805 & 35.15 & 0.9510 & 36.32 & 0.9592 & 38.75 & 0.9726 & 40.61 & 0.9799 \\ \cmidrule(l){2-16}
 & CASNet$_\text{ms;2}$ & 21.88 & 0.6012 & 26.66 & 0.7909 & 30.66 & 0.8806 & 35.46 & 0.9428 & 36.66 & 0.9524 & 38.71 & 0.9656 & 40.70 & 0.9750 \\ \cmidrule(l){2-16}
 & CASNet$_{\text{ss}}$ & 21.87 & 0.5966 & 27.22 & 0.7982 & 31.61 & 0.8893 & 36.39 & 0.9463 & 37.55 & 0.9546 & 39.75 & 0.9672 & 41.75 & 0.9757 \\ \cmidrule(l){2-16}
 & MB-RACS$_\text{ms;2}$ & 22.30 & 0.6180 & 27.41 & 0.8134 & 31.52 & 0.8986 & 36.62 & 0.9544 & 37.81 & 0.9621 & 40.00 & 0.9725 & 41.83 & 0.9790 \\ \cmidrule(l){2-16}
 & MB-RACS$_{\text{ss}}$ & \textbf{22.84} & \textbf{0.6311} & \textbf{27.93} & \textbf{0.8199} & \textbf{31.95} & \textbf{0.9064} & \textbf{37.08} & \textbf{0.9587} & \textbf{38.28} & \textbf{0.9657} & \textbf{40.42} & \textbf{0.9751} & \textbf{42.51} & \textbf{0.9815} \\ \midrule

\specialrule{0em}{1pt}{1pt}
\midrule
\multirow{11}{*}[-3.2em]{CBSD68} & CSNet & 22.37 & 0.5256 & 25.01 & 0.6603 & 27.34 & 0.7793 & 30.95 & 0.8970 & 31.96 & 0.9170 & 33.82 & 0.9444 & 35.43 & 0.9608 \\ \cmidrule(l){2-16}
 & ISTA-Net$^+$ & 22.70 & 0.5406 & 25.42 & 0.6774 & 27.92 & 0.7920 & 31.79 & 0.9038 & 32.88 & 0.9231 & 34.93 & 0.9502 & 36.88 & 0.9676 \\ \cmidrule(l){2-16}
 & Hybrid-LISTA & 22.68 & 0.5370 & 25.20 & 0.6666 & 27.57 & 0.7829 & 30.99 & 0.8961 & 32.02 & 0.9171 & 34.02 & 0.9464 & 36.03 & 0.9651 \\ \cmidrule(l){2-16}
 & SCSNet & 22.46 & 0.5320 & 25.13 & 0.6661 & 27.46 & 0.7832 & 31.08 & 0.8989 & 32.10 & 0.9186 & 33.98 & 0.9460 & 35.58 & 0.9623 \\ \cmidrule(l){2-16}
 & OPINE-Net & 22.67 & 0.5405 & 25.48 & 0.6801 & 28.01 & 0.7950 & 31.93 & 0.9055 & 33.02 & 0.9244 & 35.06 & 0.9511 & 37.08 & 0.9683 \\ \cmidrule(l){2-16}
 & AMP-Net & 22.71 & 0.5405 & 25.39 & 0.6757 & 27.89 & 0.7909 & 31.77 & 0.9031 & 32.86 & 0.9227 & 34.92 & 0.9499 & 36.93 & 0.9675 \\ \cmidrule(l){2-16}
 & COAST & 22.65 & 0.5387 & 25.42 & 0.6763 & 27.91 & 0.7910 & 31.77 & 0.9034 & 32.82 & 0.9225 & 34.84 & 0.9495 & 36.76 & 0.9667 \\ \cmidrule(l){2-16}
 & TransCS & 22.47 & 0.5213 & 25.33 & 0.6714 & 27.88 & 0.7885 & 31.87 & 0.9041 & 32.83 & 0.9215 & 35.05 & 0.9508 & 36.95 & 0.9671 \\ \cmidrule(l){2-16}
 & CASNet$_\text{ms;2}$ & 23.03 & 0.5491 & 26.00 & 0.6823 & 28.56 & 0.7892 & 32.50 & 0.8952 & 33.63 & 0.9147 & 35.82 & 0.9433 & 38.06 & 0.9628 \\ \cmidrule(l){2-16}
 & CASNet$_{\text{ss}}$ & 22.15 & 0.5456 & 25.54 & 0.6860 & 28.70 & 0.7953 & 33.75 & 0.9010 & 35.14 & 0.9200 & 37.69 & 0.9474 & 40.15 & 0.9658 \\ \cmidrule(l){2-16}
 & MB-RACS$_\text{ms;2}$ & 23.37 & 0.5617 & 26.43 & 0.7005 & 29.20 & 0.8107 & 33.51 & 0.9144 & 34.72 & 0.9319 & 37.05 & 0.9558 & 39.38 & 0.9709 \\ \cmidrule(l){2-16}
 & MB-RACS$_{\text{ss}}$ & \textbf{23.81} & \textbf{0.5719} & \textbf{26.82} & \textbf{0.7083} & \textbf{29.65} & \textbf{0.8187} & \textbf{34.08} & \textbf{0.9188} & \textbf{35.34} & \textbf{0.9352} & \textbf{37.81} & \textbf{0.9577} & \textbf{40.39} & \textbf{0.9721} \\ \midrule

 \specialrule{0em}{1pt}{1pt}
\midrule
\multirow{11}{*}[-3.2em]{Set5} & CSNet & 22.70 & 0.6016 & 27.26 & 0.7973 & 31.10 & 0.8916 & 35.73 & 0.9507 & 36.77 & 0.9588 & 38.60 & 0.9697 & 39.77 & 0.9754 \\ \cmidrule(l){2-16}
 & ISTA-Net$^+$ & 23.01 & 0.6251 & 28.02 & 0.8229 & 32.21 & 0.9063 & 36.90 & 0.9553 & 37.98 & 0.9632 & 40.05 & 0.9739 & 41.89 & 0.9814 \\ \cmidrule(l){2-16}
 & Hybrid-LISTA & 23.12 & 0.6222 & 27.69 & 0.8054 & 31.58 & 0.8949 & 35.80 & 0.9489 & 36.89 & 0.9581 & 39.00 & 0.9713 & 40.93 & 0.9796 \\ \cmidrule(l){2-16}
 & SCSNet & 22.80 & 0.6125 & 27.55 & 0.8111 & 31.35 & 0.8975 & 35.97 & 0.9524 & 36.94 & 0.9601 & 38.71 & 0.9708 & 39.78 & 0.9764 \\ \cmidrule(l){2-16}
 & OPINE-Net & 23.11 & 0.6289 & 28.20 & 0.8281 & 32.48 & 0.9104 & 37.15 & 0.9570 & 38.21 & 0.9642 & 40.27 & 0.9748 & 42.26 & 0.9819 \\ \cmidrule(l){2-16}
 & AMP-Net & 23.09 & 0.6278 & 28.00 & 0.8226 & 32.18 & 0.9064 & 36.92 & 0.9558 & 38.10 & 0.9635 & 40.14 & 0.9741 & 42.14 & 0.9816 \\ \cmidrule(l){2-16}
 & COAST & 23.06 & 0.6250 & 27.99 & 0.8208 & 32.25 & 0.9066 & 36.88 & 0.9557 & 38.00 & 0.9632 & 39.96 & 0.9738 & 41.79 & 0.9809 \\ \cmidrule(l){2-16}
 & TransCS & 23.06 & 0.6181 & 28.03 & 0.8218 & 32.44 & 0.9091 & 37.19 & 0.9576 & 38.02 & 0.9633 & 40.32 & 0.9748 & 41.95 & 0.9811 \\ \cmidrule(l){2-16}
 & CASNet$_\text{ms;2}$ & 23.46 & 0.6580 & 28.73 & 0.8327 & 33.04 & 0.9082 & 37.65 & 0.9542 & 38.78 & 0.9614 & 40.89 & 0.9719 & 43.04 & 0.9796 \\ \cmidrule(l){2-16}
 & CASNet$_{\text{ss}}$ & 22.93 & 0.6547 & 28.89 & 0.8458 & 32.52 & 0.9107 & 36.49 & 0.9523 & 39.01 & 0.9596 & 41.20 & 0.9696 & 43.18 & 0.9768 \\ \cmidrule(l){2-16}
 & MB-RACS$_\text{ms;2}$ & 23.83 & 0.6609 & 29.74 & 0.8524 & 33.73 & 0.9187 & 38.35 & 0.9594 & 39.58 & 0.9661 & 41.97 & 0.9756 & 44.20 & 0.9819 \\ \cmidrule(l){2-16}
 & MB-RACS$_{\text{ss}}$ & \textbf{24.40} & \textbf{0.6872} & \textbf{30.21} & \textbf{0.8592} & \textbf{34.03} & \textbf{0.9221} & \textbf{38.60} & \textbf{0.9617} & \textbf{39.76} & \textbf{0.9677} & \textbf{42.01} & \textbf{0.9769} & \textbf{44.34} & \textbf{0.9833} \\ \midrule

  \specialrule{0em}{1pt}{1pt}
\midrule
\multirow{11}{*}[-3.2em]{Set14} & CSNet & 21.43 & 0.5183 & 24.85 & 0.6755 & 27.76 & 0.7985 & 31.98 & 0.9059 & 32.99 & 0.9222 & 34.85 & 0.9446 & 36.19 & 0.9572 \\ \cmidrule(l){2-16}
 & ISTA-Net$^+$ & 21.84 & 0.5410 & 25.47 & 0.6974 & 28.75 & 0.8158 & 33.10 & 0.9143 & 34.27 & 0.9308 & 36.42 & 0.9525 & 38.21 & 0.9647 \\ \cmidrule(l){2-16}
 & Hybrid-LISTA & 21.95 & 0.5374 & 25.26 & 0.6845 & 28.19 & 0.8029 & 32.04 & 0.9052 & 33.05 & 0.9224 & 35.12 & 0.9471 & 37.05 & 0.9624 \\ \cmidrule(l){2-16}
 & SCSNet & 21.53 & 0.5281 & 25.04 & 0.6844 & 27.90 & 0.8029 & 32.05 & 0.9070 & 33.03 & 0.9228 & 34.81 & 0.9452 & 36.00 & 0.9577 \\ \cmidrule(l){2-16}
 & OPINE-Net & 21.88 & 0.5433 & 25.60 & 0.7033 & 28.91 & 0.8205 & 33.42 & 0.9180 & 34.62 & 0.9335 & 36.70 & 0.9538 & 38.64 & \textbf{0.9667} \\ \cmidrule(l){2-16}
 & AMP-Net & 21.92 & 0.5430 & 25.53 & 0.6980 & 28.77 & 0.8154 & 33.18 & 0.9142 & 34.32 & 0.9298 & 36.55 & 0.9525 & 38.52 & 0.9661 \\ \cmidrule(l){2-16}
 & COAST & 21.83 & 0.5402 & 25.59 & 0.6994 & 28.73 & 0.8156 & 33.14 & 0.9144 & 34.24 & 0.9298 & 36.29 & 0.9514 & 38.11 & 0.9651 \\ \cmidrule(l){2-16}
 & TransCS & 21.73 & 0.5154 & 25.58 & 0.6958 & 28.78 & 0.8152 & 33.35 & 0.9164 & 34.43 & 0.9305 & 36.79 & 0.9536 & 38.42 & 0.9657 \\ \cmidrule(l){2-16}
 & CASNet$_\text{ms;2}$ & 22.50 & 0.5612 & 26.18 & 0.7043 & 29.19 & 0.8108 & 33.56 & 0.9052 & 34.63 & 0.9209 & 36.68 & 0.9442 & 38.74 & 0.9604 \\ \cmidrule(l){2-16}
 & CASNet$_{\text{ss}}$ & 22.26 & 0.5524 & 26.40 & 0.7095 & 29.77 & 0.8190 & 34.68 & 0.9055 & 35.77 & 0.9193 & 37.73 & 0.9401 & 39.59 & 0.9556 \\ \cmidrule(l){2-16}
 & MB-RACS$_\text{ms;2}$ & 22.86 & 0.5757 & 26.84 & 0.7252 & 30.21 & 0.8323 & 34.52 & 0.9189 & 35.63 & 0.9330 & 37.76 & 0.9524 & 39.67 & 0.9645 \\ \cmidrule(l){2-16}
 & MB-RACS$_{\text{ss}}$ & \textbf{23.51} & \textbf{0.5872} & \textbf{27.37} & \textbf{0.7325} & \textbf{30.72} & \textbf{0.8390} & \textbf{35.16} & \textbf{0.9239} & \textbf{36.27} & \textbf{0.9366} & \textbf{38.31} & \textbf{0.9541} & \textbf{40.24} & 0.9655 \\ \midrule

 \specialrule{0em}{4pt}{1pt}
 \bottomrule
 \end{tabular}
}
\end{table*}

%% file: Section/experiments_and_results_section/tables/single-stage_table_2_datasets_different_methods_results.tex
\begin{table*}
\centering
\caption{PSNR and SSIM comparisons for MB-RACS and 9 baselines on BSDS100, Urban100 and Kodak24 datasets at sampling rates of 0.01, 0.04, 0.1, 0.25, 0.3, 0.4, and 0.5. The subscript ``ss'' denotes the use of the single-stage rate-adaptive sampling method. The subscript ``ms;$x$'' denotes the use of the $x$-stage rate-adaptive sampling method.}
\label{tab:comparison with state-of-the-arts methods on two datasets}
\resizebox{\linewidth}{!}{
\begin{tabular}{c c c c c c c c c c c c c c c c}
\toprule
\multirow{2}{*}[-1em]{Dataset} & \multirow{2}{*}[-1em]{Method} & \multicolumn{14}{c}{Sampling Ratio} \\ \cmidrule(lr){3-16} 
& & \multicolumn{2}{c}{0.01} & \multicolumn{2}{c}{0.04} & \multicolumn{2}{c}{0.1} & \multicolumn{2}{c}{0.25} & \multicolumn{2}{c}{0.3} & \multicolumn{2}{c}{0.4} & \multicolumn{2}{c}{0.5} \\
\cmidrule(lr){3-4} \cmidrule(lr){5-6} \cmidrule(lr){7-8} \cmidrule(lr){9-10} \cmidrule(lr){11-12} \cmidrule(lr){13-14} \cmidrule(lr){15-16} & & PSNR & SSIM & PSNR & SSIM & PSNR & SSIM & PSNR & SSIM & PSNR & SSIM & PSNR & SSIM & PSNR & SSIM \\
\midrule

\specialrule{0em}{1pt}{1pt}
\midrule
\multirow{11}{*}[-3.2em]{BSDS100} & CSNet & 22.34 & 0.5095 & 24.88 & 0.6441 & 27.13 & 0.7654 & 30.70 & 0.8893 & 31.70 & 0.9107 & 33.57 & 0.9402 & 35.20 & 0.9579 \\ \cmidrule(l){2-16}
 & ISTA-Net$^+$ & 22.65 & 0.5246 & 25.25 & 0.6606 & 27.68 & 0.7779 & 31.50 & 0.8959 & 32.58 & 0.9166 & 34.64 & 0.9459 & 36.59 & 0.9647 \\ \cmidrule(l){2-16}
 & Hybrid-LISTA & 22.65 & 0.5211 & 25.06 & 0.6503 & 27.34 & 0.7690 & 30.72 & 0.8883 & 31.75 & 0.9107 & 33.76 & 0.9421 & 35.77 & 0.9623 \\ \cmidrule(l){2-16}
 & SCSNet & 22.43 & 0.5157 & 24.99 & 0.6497 & 27.24 & 0.7693 & 30.82 & 0.8912 & 31.83 & 0.9123 & 33.72 & 0.9418 & 35.34 & 0.9594 \\ \cmidrule(l){2-16}
 & OPINE-Net & 22.63 & 0.5245 & 25.31 & 0.6634 & 27.77 & 0.7810 & 31.64 & 0.8978 & 32.72 & 0.9180 & 34.77 & 0.9468 & 36.78 & 0.9654 \\ \cmidrule(l){2-16}
 & AMP-Net & 22.67 & 0.5243 & 25.22 & 0.6589 & 27.64 & 0.7766 & 31.48 & 0.8952 & 32.56 & 0.9162 & 34.63 & 0.9456 & 36.64 & 0.9645 \\ \cmidrule(l){2-16}
 & COAST & 22.61 & 0.5225 & 25.24 & 0.6595 & 27.66 & 0.7767 & 31.47 & 0.8954 & 32.52 & 0.9160 & 34.54 & 0.9451 & 36.46 & 0.9638 \\ \cmidrule(l){2-16}
 & TransCS & 22.44 & 0.5056 & 25.17 & 0.6547 & 27.63 & 0.7741 & 31.58 & 0.8963 & 32.52 & 0.9147 & 34.76 & 0.9465 & 36.68 & 0.9643 \\ \cmidrule(l){2-16}
 & CASNet$_\text{ms;2}$ & 22.98 & 0.5335 & 25.80 & 0.6657 & 28.36 & 0.7764 & 32.27 & 0.8880 & 33.38 & 0.9085 & 35.56 & 0.9390 & 37.77 & 0.9597 \\ \cmidrule(l){2-16}
 & CASNet$_{\text{ss}}$ & 22.26 & 0.5290 & 25.47 & 0.6688 & 28.55 & 0.7811 & 33.44 & 0.8923 & 34.76 & 0.9125 & 37.26 & 0.9420 & 39.69 & 0.9618 \\ \cmidrule(l){2-16}
 & MB-RACS$_\text{ms;2}$ & 23.29 & 0.5454 & 26.20 & 0.6827 & 28.91 & 0.7963 & 33.17 & 0.9063 & 34.37 & 0.9251 & 36.68 & 0.9512 & 38.99 & 0.9678 \\ \cmidrule(l){2-16}
 & MB-RACS$_{\text{ss}}$ & \textbf{23.68} & \textbf{0.5544} & \textbf{26.59} & \textbf{0.6903} & \textbf{29.34} & \textbf{0.8037} & \textbf{33.69} & \textbf{0.9102} & \textbf{34.92} & \textbf{0.9280} & \textbf{37.35} & \textbf{0.9526} & \textbf{39.87} & \textbf{0.9686} \\ \midrule

\specialrule{0em}{1pt}{1pt} 
\midrule
\multirow{10}{*}[-2.8em]{Urban100} & CSNet & 19.31 & 0.4766 & 22.26 & 0.6492 & 25.00 & 0.7848 & 29.12 & 0.8999 & 30.18 & 0.9178 & 32.09 & 0.9420 & 33.71 & 0.9573 \\ \cmidrule(l){2-16}
 & ISTA-Net$^+$ & 19.65 & 0.5008 & 23.02 & 0.6905 & 26.39 & 0.8234 & 31.25 & 0.9227 & 32.57 & 0.9396 & 34.82 & 0.9601 & 36.60 & 0.9723 \\ \cmidrule(l){2-16}
 & Hybrid-LISTA & 19.72 & 0.4958 & 22.61 & 0.6651 & 25.42 & 0.7931 & 29.09 & 0.8964 & 30.20 & 0.9167 & 32.31 & 0.9448 & 34.41 & 0.9629 \\ \cmidrule(l){2-16}
 & SCSNet & 19.39 & 0.4819 & 22.45 & 0.6585 & 25.24 & 0.7906 & 29.39 & 0.9008 & 30.46 & 0.9183 & 32.35 & 0.9415 & 33.89 & 0.9556 \\ \cmidrule(l){2-16}
 & OPINE-Net & 19.63 & 0.5030 & 23.20 & 0.7019 & 26.75 & 0.8341 & 31.84 & 0.9303 & 33.09 & 0.9444 & 35.27 & 0.9630 & 37.26 & 0.9748 \\ \cmidrule(l){2-16}
 & AMP-Net & 19.70 & 0.4998 & 22.95 & 0.6874 & 26.32 & 0.8210 & 31.34 & 0.9240 & 32.61 & 0.9396 & 34.86 & 0.9602 & 36.91 & 0.9732 \\ \cmidrule(l){2-16}
 & COAST & 19.65 & 0.5019 & 23.08 & 0.6915 & 26.43 & 0.8229 & 31.24 & 0.9238 & 32.42 & 0.9386 & 34.55 & 0.9590 & 36.40 & 0.9716 \\ \cmidrule(l){2-16}
 & CASNet$_\text{ms;2}$ & 20.15 & 0.5320 & 23.85 & 0.7154 & 27.14 & 0.8272 & 31.67 & 0.9157 & 32.84 & 0.9304 & 35.04 & 0.9519 & 37.19 & 0.9666 \\ \cmidrule(l){2-16}
 & CASNet$_{\text{ss}}$ & 19.21 & 0.5228 & 23.26 & 0.7177 & 27.26 & 0.8313 & 33.18 & 0.9196 & 34.50 & 0.9338 & 36.91 & 0.9544 & 39.26 & 0.9684 \\ \cmidrule(l){2-16}
 & MB-RACS$_\text{ms;2}$ & 20.66 & 0.5565 & 24.68 & 0.7459 & 28.28 & 0.8556 & 33.10 & 0.9352 & 34.30 & 0.9472 & 36.48 & 0.9635 & 38.62 & 0.9743 \\ \cmidrule(l){2-16}
 & MB-RACS$_{\text{ss}}$ & \textbf{21.12} & \textbf{0.5713} & \textbf{25.21} & \textbf{0.7589} & \textbf{28.92} & \textbf{0.8657} & \textbf{33.86} & \textbf{0.9396} & \textbf{35.10} & \textbf{0.9505} & \textbf{37.47} & \textbf{0.9655} & \textbf{39.84} & \textbf{0.9755} \\ \midrule

\specialrule{0em}{1pt}{1pt}
\midrule
\multirow{11}{*}[-3.2em]{Kodak24} & CSNet & 23.31 & 0.5719 & 25.80 & 0.6912 & 28.17 & 0.7999 & 32.04 & 0.9088 & 33.12 & 0.9266 & 35.17 & 0.9508 & 36.90 & 0.9651 \\ \cmidrule(l){2-16}
 & ISTA-Net$^+$ & 23.65 & 0.5874 & 26.35 & 0.7141 & 29.03 & 0.8205 & 33.32 & 0.9200 & 34.53 & 0.9368 & 36.79 & 0.9593 & 38.85 & 0.9735 \\ \cmidrule(l){2-16}
 & Hybrid-LISTA & 23.65 & 0.5846 & 26.04 & 0.6987 & 28.48 & 0.8056 & 32.12 & 0.9084 & 33.23 & 0.9273 & 35.44 & 0.9536 & 37.64 & 0.9703 \\ \cmidrule(l){2-16}
 & SCSNet & 23.41 & 0.5778 & 25.91 & 0.6971 & 28.30 & 0.8044 & 32.22 & 0.9110 & 33.30 & 0.9285 & 35.34 & 0.9526 & 37.05 & 0.9666 \\ \cmidrule(l){2-16}
 & OPINE-Net & 23.67 & 0.5889 & 26.48 & 0.7193 & 29.18 & 0.8249 & 33.57 & 0.9228 & 34.78 & 0.9388 & 37.02 & 0.9606 & 39.17 & 0.9743 \\ \cmidrule(l){2-16}
 & AMP-Net & 23.67 & 0.5869 & 26.28 & 0.7103 & 28.93 & 0.8177 & 33.30 & 0.9190 & 34.50 & 0.9360 & 36.78 & 0.9592 & 38.95 & 0.9735 \\ \cmidrule(l){2-16}
 & COAST & 23.61 & 0.5855 & 26.31 & 0.7177 & 28.99 & 0.8186 & 33.29 & 0.9193 & 34.45 & 0.9357 & 36.65 & 0.9585 & 38.73 & 0.9728 \\ \cmidrule(l){2-16}
 & TransCS & 23.37 & 0.5580 & 26.32 & 0.7098 & 29.02 & 0.8187 & 33.47 & 0.9205 & 34.56 & 0.9354 & 37.00 & 0.9601 & 39.00 & 0.9726 \\ \cmidrule(l){2-16}
 & CASNet$_\text{ms;2}$ & 24.13 & 0.6009 & 27.21 & 0.7244 & 30.13 & 0.8191 & 34.55 & 0.9126 & 35.71 & 0.9291 & 37.91 & 0.9533 & 40.06 & 0.9692 \\ \cmidrule(l){2-16}
 & CASNet$_{\text{ss}}$ & 23.06 & 0.5996 & 26.72 & 0.7282 & 30.89 & 0.8244 & 35.62 & 0.9133 & 36.92 & 0.9294 & 39.28 & 0.9529 & 41.47 & 0.9684 \\ \cmidrule(l){2-16}
 & MB-RACS$_\text{ms;2}$ & 24.57 & 0.6155 & 27.77 & 0.7444 & 30.88 & 0.8405 & 35.54 & 0.9276 & 36.79 & 0.9419 & 39.06 & 0.9611 & 41.19 & 0.9731 \\ \cmidrule(l){2-16}
 & MB-RACS$_{\text{ss}}$ & \textbf{25.08} & \textbf{0.6278} & \textbf{28.32} & \textbf{0.7553} & \textbf{31.35} & \textbf{0.8498} & \textbf{35.99} & \textbf{0.9328} & \textbf{37.27} & \textbf{0.9459} & \textbf{39.68} & \textbf{0.9635} & \textbf{41.97} & \textbf{0.9744} \\ \midrule
 \specialrule{0em}{4pt}{1pt}
 \bottomrule
 \end{tabular}
} 
\end{table*}

%% file: Section/proposed_method_subsection/cross-iteration_skip_connections.tex
\subsection{Decoder-Side Cross-Iteration Skip Connections}\label{sec:3:cross-iteration skip connections}

When decoding the sampled signals, we first perform an initial reconstruction for each block:
\begin{align}
    \hat{\textit{\textbf{x}}}_i^0 = {\textbf{\textit{A}}_{1:M_i}}^{\transpose} \textit{\textbf{y}}_i. 
\end{align}

\input{Section/experiments_and_results_section/figures/single-stage_figure_different_methods_visual_results_sampling_rate_0.1}
\input{Section/experiments_and_results_section/figures/single-stage_figure_different_methods_visual_results_sampling_rate_0.01}

Next, the reconstructions of each block are concatenated to form the complete image's initial reconstruction, denoted as $\hat{\textit{\textbf{x}}}^0$. Utilizing the iterative decoder proposed in \cite{9854112}, the reconstruction starts with $\hat{\textit{\textbf{x}}}^0$ and undergoes $L$ iterations to yield the reconstructed image. Inspired by the validation in \cite{huang2017densely} that dense skip connections can strengthen feature propagation and encourage feature reuse, we propose cross-iteration skip connections added to the aforementioned decoding structure to promote feature information propagation across iterations. The structure of the cross-iteration skip connections is shown in Fig.~\ref{fig:cross-iteration_skip_connection}. The cross-iteration skip connections indicates that the output features of the ${l-1}^\text{th}$ iteration U-net's three upsampling blocks and body block are added to the features in the ${l}^\text{th}$ iteration U-net's three downsampling blocks and body block, respectively. 

After obtaining the final reconstruction $\hat{\textit{\textbf{x}}}^L$, we train MB-RACS using the MSE loss between the reconstructed image and the original image, combined with the loss mentioned previously in \eqref{eq:KL loss for measurement bounds prediction network training}:
\begin{align}
    \mathcal{L}_{\text{total}} = \gamma\frac{1}{HW} \|\hat{\textit{\textbf{x}}}^L-\textit{\textbf{x}}\|^2_2 + \mathcal{L}_{\text{KL}},
\end{align}
where $\gamma$ is the weight for the MSE loss. The MSE loss is used to train the measurement matrix $\textit{\textbf{A}}$ and the reconstruction end's network parameters. As mentioned in Sec.~\ref{sec:3:multi-stage rate-adaptive sampling strategy:sampling process:sampling in subsequent stages}, $\mathcal{L}_{\text{KL}}$ is used to train the parameters of the measurement bounds prediction network.

%% file: Section/experiments_and_results_section/figures/single-stage_figure_different_methods_visual_results_sampling_rate_0.1.tex
\begin{figure*}
    \centering
    \newlength{\mylength}
    \setlength{\mylength}{\dimexpr\textwidth*100/901}
    \begin{minipage}[t]{\mylength}
		\centering
            \centerline{\footnotesize{Ground Truth}}	
            \vspace{3.6pt}
		\includegraphics[width=\linewidth]                    {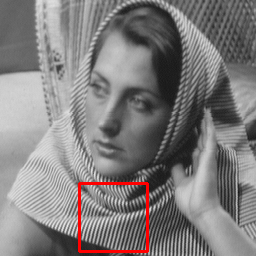}
            \includegraphics[width=\linewidth]{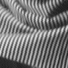}

            \vspace{-5pt}
		\centerline{\footnotesize{PSNR/SSIM}}	
            \vspace{4pt}
            \includegraphics[width=\linewidth]                    {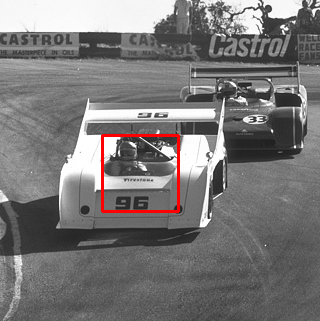}
            \includegraphics[width=\linewidth]{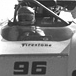}
            
            \vspace{-5pt}
		\centerline{\footnotesize{PSNR/SSIM}}
    \end{minipage}%
    \hfill
    \begin{minipage}[t]{\mylength}
		\centering
            \centerline{\footnotesize{CSNet}}
            \vspace{3.6pt}
		\includegraphics[width=\linewidth]{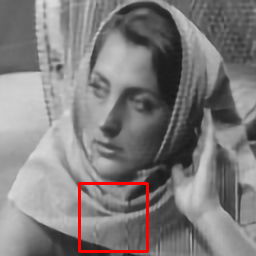}
            \includegraphics[width=\linewidth]{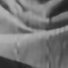}
            
            \vspace{-5pt}
		\centerline{\footnotesize{24.37/0.7212}}	
            \vspace{4pt}
            \includegraphics[width=\linewidth]                    {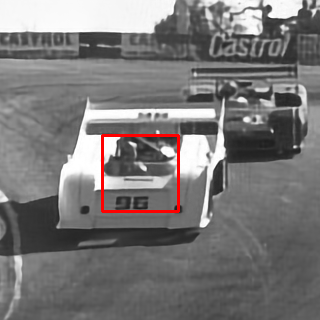}
            \includegraphics[width=\linewidth]{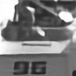}
            
            \vspace{-5pt}
		\centerline{\footnotesize{26.94/0.7737}}
    \end{minipage}%
    \hfill
    \begin{minipage}[t]{\mylength}
		\centering
            \centerline{\footnotesize{ISTA-Net$^+$}}
            \vspace{3.6pt}
		\includegraphics[width=\linewidth]{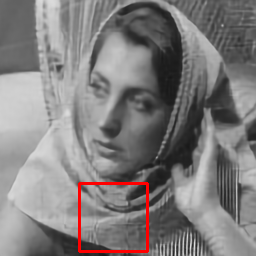}
            \includegraphics[width=\linewidth]{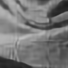}
            
            \vspace{-5pt}
		\centerline{\footnotesize{25.05/0.7492}}
            \vspace{4pt}
            \includegraphics[width=\linewidth]                    {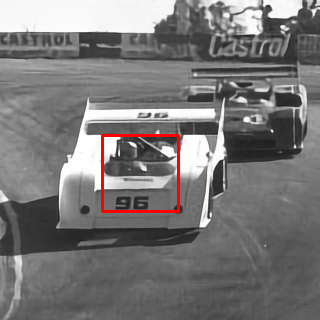}
            \includegraphics[width=\linewidth]{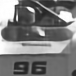}
            
            \vspace{-5pt}
		\centerline{\footnotesize{27.93/0.7945}}
    \end{minipage}%
    \hfill
    \begin{minipage}[t]{\mylength}
		\centering
            \centerline{\footnotesize{SCSNet}}
            \vspace{3.6pt}
		\includegraphics[width=\linewidth]{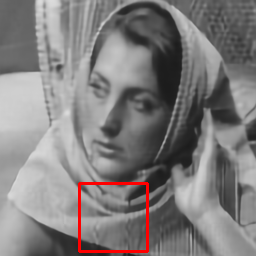}
            \includegraphics[width=\linewidth]{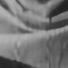}
            
            \vspace{-5pt}
		\centerline{\footnotesize{24.38/0.7227}}
            \vspace{4pt}
            \includegraphics[width=\linewidth]                    {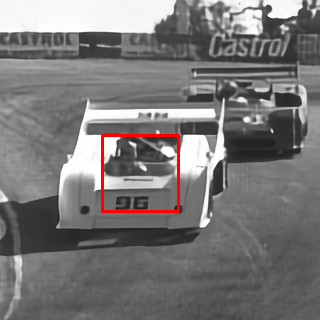}
            \includegraphics[width=\linewidth]{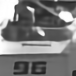}
            
            \vspace{-5pt}
		\centerline{\footnotesize{27.12/0.7791}}
    \end{minipage}%
    \hfill
    \begin{minipage}[t]{\mylength}
		\centering
            \centerline{\footnotesize{OPINE-Net}}
            \vspace{3.6pt}
		\includegraphics[width=\linewidth]{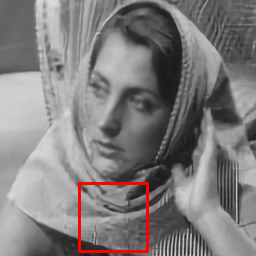}
            \includegraphics[width=\linewidth]{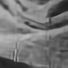}
            
            \vspace{-5pt}
		\centerline{\footnotesize{25.12/0.7531}}	
            \vspace{4pt}
            \includegraphics[width=\linewidth]                    {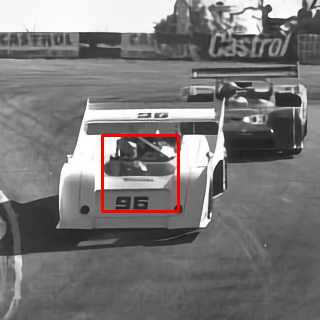}
            \includegraphics[width=\linewidth]{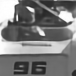}
            
            \vspace{-5pt}
		\centerline{\footnotesize{28.09/0.7993}}
    \end{minipage}%
    \hfill
    \begin{minipage}[t]{\mylength}
		\centering
            \centerline{\footnotesize{AMP-Net}}
            \vspace{3.6pt}
		\includegraphics[width=\linewidth]{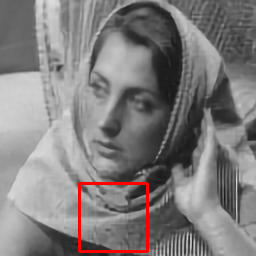}
            \includegraphics[width=\linewidth]{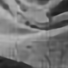}
            
            \vspace{-5pt}
		\centerline{\footnotesize{25.00/0.7455}}
            \vspace{4pt}
            \includegraphics[width=\linewidth]                    {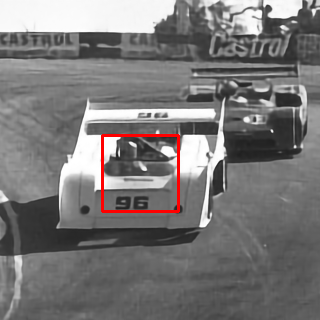}
            \includegraphics[width=\linewidth]{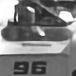}
            
            \vspace{-5pt}
		\centerline{\footnotesize{27.81/0.7924}}
    \end{minipage}%
    \hfill
    \begin{minipage}[t]{\mylength}
		\centering
            \centerline{\footnotesize{COAST}}
            \vspace{3.6pt}
		\includegraphics[width=\linewidth]{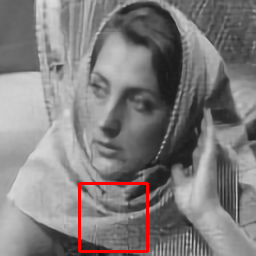}
            \includegraphics[width=\linewidth]{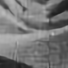}
            
            \vspace{-5pt}
		\centerline{\footnotesize{24.98/0.7445}}	
            \vspace{4pt}
            \includegraphics[width=\linewidth]                    {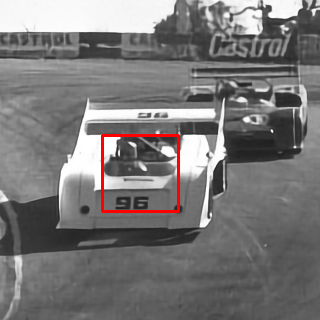}
            \includegraphics[width=\linewidth]{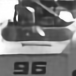}
            
            \vspace{-5pt}
		\centerline{\footnotesize{27.90/0.7922}}
    \end{minipage}%
    \hfill
    \begin{minipage}[t]{\mylength}
		\centering
            \centerline{\footnotesize{CASNet$_\text{ms;2}$}}
            \vspace{1.5pt}
		\includegraphics[width=\linewidth]{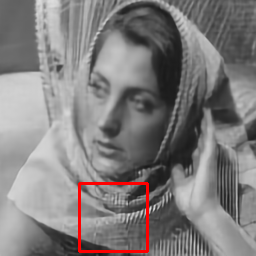}
            \includegraphics[width=\linewidth]{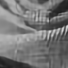}
            
            \vspace{-5pt}
		\centerline{\footnotesize{25.32/0.7597}}	
            \vspace{4pt}
            \includegraphics[width=\linewidth]{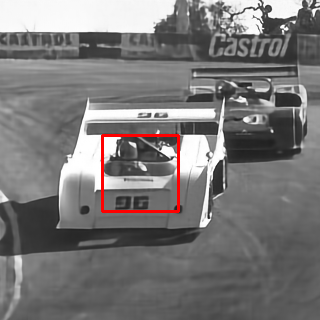}
            \includegraphics[width=\linewidth]{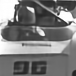}
            
            \vspace{-5pt}
		\centerline{\footnotesize{29.02/0.7825}}
    \end{minipage}%
    \hfill
    \begin{minipage}[t]{\mylength}
		\centering
            \centerline{\footnotesize{MB-RACS$_\text{ms;2}$}}
            \vspace{1.5pt}
		\includegraphics[width=\linewidth]{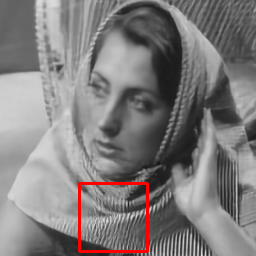}
            \includegraphics[width=\linewidth]{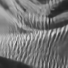}
            
            \vspace{-5pt}
		\centerline{\footnotesize{26.24/0.7918}}	
            \vspace{4pt}
            \includegraphics[width=\linewidth]{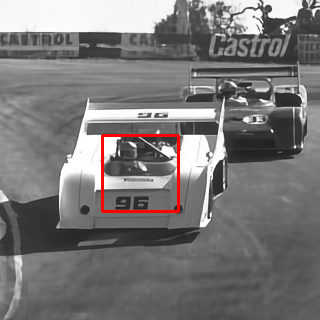}
            \includegraphics[width=\linewidth]{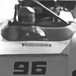}
            
            \vspace{-5pt}
		\centerline{\footnotesize{29.70/0.8063}}
    \end{minipage}%
    \caption{Visual comparison of two images from Set11 (top) and CBSD68 (bottom) using different methods at a 0.1 sampling rate.}
    \label{fig:visual comparison 1}
\end{figure*}

%% file: Section/experiments_and_results_section/figures/single-stage_figure_different_methods_visual_results_sampling_rate_0.01.tex
\begin{figure}
    \centering
    \newlength{\mylengthb}
    \setlength{\mylengthb}{\dimexpr18.5\linewidth/50}
    \newlength{\mylengthc}
    \setlength{\mylengthc}{\dimexpr10.7\linewidth/70}
    \begin{minipage}[t]{\mylengthb}
            \vspace{0pt}
		\centering
		\includegraphics[width=\linewidth]{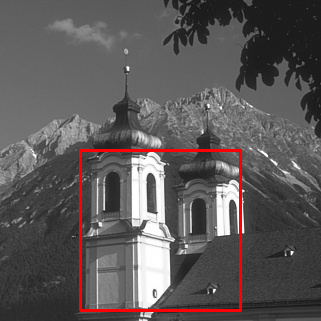}
  
            \vspace{-5pt}
            
		\centerline{\scriptsize{PSNR/SSIM}}	
            \vspace{-5pt}
            \centerline{\scriptsize{Groud Truth}}		
    \end{minipage}%
    \hfill
    \begin{minipage}[t]{\mylengthc}
            \vspace{0pt}
		\centering
		\includegraphics[width=\linewidth]{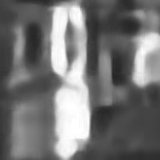}

            \vspace{-6pt}
            
		\centerline{\scriptsize{23.5/0.657}}	
            \vspace{-5pt}	
            \centerline{\scriptsize{CSNet}}	
            \vspace{3pt}	
            \includegraphics[width=\linewidth]{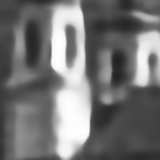}

            \vspace{-5pt}
            
		\centerline{\scriptsize{24.1/0.670}}	
            \vspace{-5pt}
            \centerline{\scriptsize{AMP-Net}}	
    \end{minipage}%
    \hfill
    \begin{minipage}[t]{\mylengthc}
            \vspace{0pt}
		\centering
		\includegraphics[width=\linewidth]{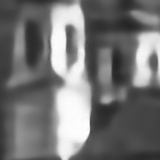}

            \vspace{-6pt}
            
		\centerline{\scriptsize{24.1/0.670}}	
            \vspace{-5pt}	
            \centerline{\scriptsize{ISTA-Net$^+$}}	
            \vspace{3pt}	
            \includegraphics[width=\linewidth]{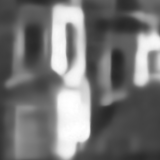}

            \vspace{-5pt}
            
		\centerline{\scriptsize{24.0/0.668}}	
            \vspace{-5pt}
            \centerline{\scriptsize{COAST}}	
    \end{minipage}%
    \hfill
    \begin{minipage}[t]{\mylengthc}
            \vspace{0pt}
		\centering
		\includegraphics[width=\linewidth]{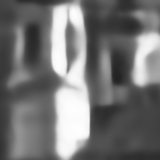}

            \vspace{-6pt}
            
		\centerline{\scriptsize{23.6/0.664}}	
            \vspace{-5pt}	
            \centerline{\scriptsize{SCSNet}}	
            \vspace{3pt}	
            \includegraphics[width=\linewidth]{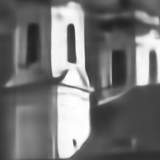}

            \vspace{-5pt}
            
		\centerline{\scriptsize{25.5/0.703}}	
            \vspace{-5pt}
            \centerline{\scriptsize{CASNet$_\text{ss}$}}	
    \end{minipage}%
    \hfill
    \begin{minipage}[t]{\mylengthc}
            \vspace{0pt}
		\centering
		\includegraphics[width=\linewidth]{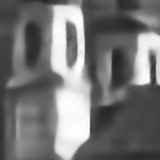}

            \vspace{-6pt}
            
		\centerline{\scriptsize{24.2/0.673}}	
            \vspace{-5pt}	
            \centerline{\scriptsize{OPINE-Net}}	
            \vspace{3pt}
            \includegraphics[width=\linewidth]{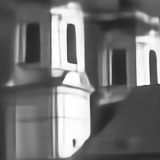}

            \vspace{-5pt}
            
		\centerline{\scriptsize{26.1/0.715}}	
            \vspace{-5pt}
            \centerline{\scriptsize{MB-RACS$_\text{ss}$}}	
    \end{minipage}%
    \hfill
    \vspace{7pt}
    \begin{minipage}[t]{\mylengthb}
            \vspace{0pt}
		\centering
		\includegraphics[width=\linewidth]{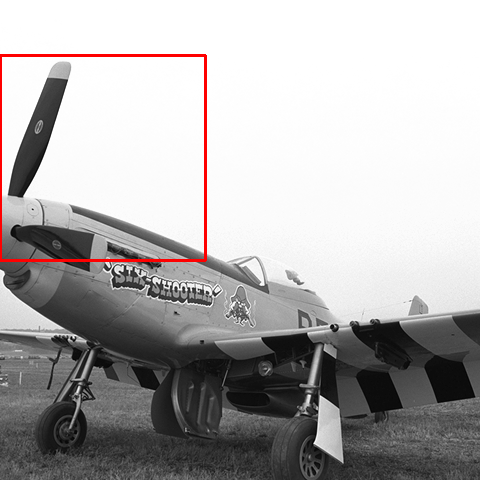}
  
            \vspace{-5pt}
            
		\centerline{\scriptsize{PSNR/SSIM}}	
            \vspace{-5pt}
            \centerline{\scriptsize{Groud Truth}}		
    \end{minipage}%
    \hfill
    \begin{minipage}[t]{\mylengthc}
            \vspace{0pt}
		\centering
		\includegraphics[width=\linewidth]{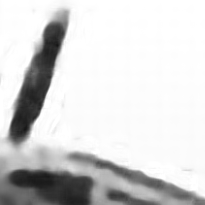}

            \vspace{-6pt}
            
		\centerline{\scriptsize{24.3/0.765}}	
            \vspace{-5pt}	
            \centerline{\scriptsize{CSNet}}	
            \vspace{3pt}	
            \includegraphics[width=\linewidth]{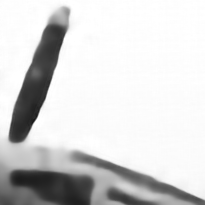}

            \vspace{-5pt}
            
		\centerline{\scriptsize{25.1/0.783}}	
            \vspace{-5pt}
            \centerline{\scriptsize{AMP-Net}}	
    \end{minipage}%
    \hfill
    \begin{minipage}[t]{\mylengthc}
            \vspace{0pt}
		\centering
		\includegraphics[width=\linewidth]{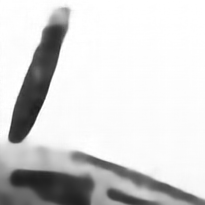}

            \vspace{-6pt}
            
		\centerline{\scriptsize{25.1/0.785}}	
            \vspace{-5pt}	
            \centerline{\scriptsize{ISTA-Net$^+$}}	
            \vspace{3pt}	
            \includegraphics[width=\linewidth]{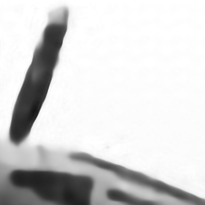}

            \vspace{-5pt}
            
		\centerline{\scriptsize{25.0/0.780}}	
            \vspace{-5pt}
            \centerline{\scriptsize{COAST}}	
    \end{minipage}%
    \hfill
    \begin{minipage}[t]{\mylengthc}
            \vspace{0pt}
		\centering
		\includegraphics[width=\linewidth]{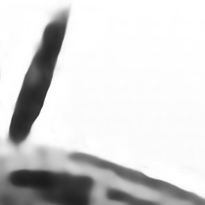}

            \vspace{-6pt}
            
		\centerline{\scriptsize{24.6/0.775}}	
            \vspace{-5pt}	
            \centerline{\scriptsize{SCSNet}}	
            \vspace{3pt}	
            \includegraphics[width=\linewidth]{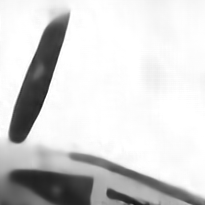}

            \vspace{-5pt}
            
		\centerline{\scriptsize{22.1/0.801}}	
            \vspace{-5pt}
            \centerline{\scriptsize{CASNet$_\text{ss}$}}	
    \end{minipage}%
    \hfill
    \begin{minipage}[t]{\mylengthc}
            \vspace{0pt}
		\centering
		\includegraphics[width=\linewidth]{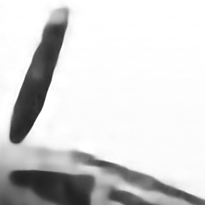}

            \vspace{-6pt}
            
		\centerline{\scriptsize{24.9/0.784}}	
            \vspace{-5pt}	
            \centerline{\scriptsize{OPINE-Net}}	
            \vspace{3pt}
            \includegraphics[width=\linewidth]{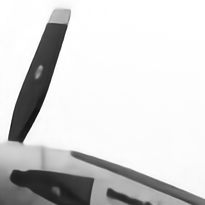}

            \vspace{-5pt}
            
		\centerline{\scriptsize{27.4/0.831}}	
            \vspace{-5pt}
            \centerline{\scriptsize{MB-RACS$_\text{ss}$}}	
    \end{minipage}%
    \caption{Visual comparison of two images from BSDS100 (top) and Kodak24 (bottom) using different methods at a 0.01 sampling rate.}
    \label{fig:visual comparison 2}
\end{figure}

%% file: Section/experiments_and_results_section/experiments_and_results.tex
\section{Experiments and Results}\label{sec:experiment_and_result}

\subsection{Experimental Setup}
\label{sec:4:experimental setup}

\subsubsection{Datasets}
\label{sec:4:experimental setup:datasets}
Our training set aligns with \cite{8765626}. On 400 images from the training and test datasets of the BSDS500 database \cite{5557884}, we applied 8 data augmentation techniques. These images were then cropped into 96 × 96 sub-images, yielding a training set of 89,600 sub-images. Moreover, we conducted test experiments on seven datasets: Set11 \cite{Kulkarni_2016_CVPR}, CBSD68 \cite{937655}, Set5 \cite{bevilacqua2012low}, Set14 \cite{zeyde2012single}, BSDS100 \cite{937655}, Urban100 \cite{huang2015single}, and Kodak24 \cite{franzen1999kodak}. These datasets are commonly used as test sets in image compressive sensing and other image processing fields, containing images that range in number from a few to 100.

\subsubsection{Evaluation Metrics}
\label{sec:4:experimental setup:metrics}
We utilize two primary metrics to assess image quality and signal distortion in our experiments. Specifically, the peak signal-to-noise ratio (PSNR) provides a standard measure of reconstruction fidelity, while the multi-scale structural similarity (MS-SSIM) \cite{wang2003multiscale} offers insights into perceived image quality across different scales.

\subsubsection{Baselines}
\label{sec:4:experimental setup:baselines}

We compare the MB-RACS with 8 baselines: CSNet \cite{8765626}, ISTA-Net$^+$ \cite{Zhang_2018_CVPR}, Hybrid-LISTA \cite{9767634}, SCSNet \cite{Shi_2019_CVPR}, OPINE-Net \cite{9019857}, AMP-Net \cite{9298950}, COAST \cite{9467810}, TransCS \cite{9934025}, and CASNet \cite{9854112}. 
CSNet is a purely network-based method. ISTA-Net$^+$, Hybrid-LISTA, OPINE-Net, AMP-Net, and TransCS are deep-unfolding-based methods. SCSNet achieves a scalable sampling rate through a scalable convolutional neural network. COAST allows sampling to generalize to any measurement matrix. CASNet designs a rate-adaptive sampling method based on the saliency map.

\subsubsection{Implementation Details}
\label{sec:4:experimental setup:implementation_details}

The single-stage rate-adaptive sampling model trains for 150 epochs, while the multi-stage rate-adaptive sampling model runs for 120 epochs. The learning rate starts at \(1 \times 10^{-4}\) for the initial 100 epochs and then decreases to \(1 \times 10^{-5}\) thereafter. The Adam optimizer is employed, with coefficients 0.9 and 0.999 used to compute running averages of the gradient and its square, respectively.
Other hyper-parameters are listed as follows: 
(\textbf{i}) The patch size, $B=32$.
(\textbf{ii}) The parameters for fitting the pair $(s_\text{r}, p_\text{s})$, $a=78.77,\ b=0.0444,\ s_{r;1}=0.01,\ p_{s;1}=0.005$.
(\textbf{iii}) The number of iterations at the decoder side, $L=5$.
(\textbf{iv}) The weight for the MSE loss, $\gamma=200$.

\input{Section/experiments_and_results_section/figures/two-stage_figure_two_methods_visual_and_ratio_allocation_results_sampling_rate_0.04}

\input{Section/experiments_and_results_section/figures/single-stage_figure_two_methods_visual_and_ratio_allocation_results_sampling_rate_0.25}

\subsection{Results and Analysis}
\label{sec:4:result and analysis}

\subsubsection{Comparison with State-of-the-Arts Methods}
\label{sec:4:result and analysis:comparison with state-of-the-arts methods}

Tab. \ref{tab:comparison with state-of-the-arts methods on five datasets} and Tab. \ref{tab:comparison with state-of-the-arts methods on two datasets} present the PSNR and SSIM results of MB-RACS compared with 9 baselines across seven datasets at seven different sampling rates. We not only compare with the single-stage rate-adaptive version of CASNet but also with its two-stage version. In this two-stage version, the first stage employs uniform sampling, whereas the second stage uses the data from the first stage to infer a saliency map and implement rate-adaptive sampling. For labeling, the subscript "ss" is used with both CASNet and MB-RACS to denote single-stage rate-adaptive sampling. The subscript "ms;$x$" is used to denote multi-stage rate-adaptive sampling, where $x$ indicates the number of sampling stages. When comparing MB-RACS$_{\text{ss}}$ with uniform sampling methods such as CSNet, ISTA-Net$^+$, Hybrid-LISTA, SCSNet, OPINE-Net, AMP-Net, COAST, and TransCS, MB-RACS consistently demonstrates advantages across various datasets. For example, it exhibits an average gain of 1.93dB/0.0175 against OPINE-Net and 2.12dB/0.0208 against AMP-Net in PSNR/SSIM metrics. Compared to the single-stage rate-adaptive sampling version of CASNet, MB-RACS$_{\text{ss}}$ also demonstrates an average gain of 0.85dB/0.0182. Moreover, MB-RACS$_\text{ms;2}$ also outperforms uniform sampling methods. For example, compared to OPINE-Net, it achieves an average improvement of 1.38dB/0.0119. When compared to the two-stage rate-adaptive sampling version of CASNet, MB-RACS$_\text{ms;2}$ also exhibits an average gain of 0.90dB/0.0141. Fig.~\ref{fig:visual comparison 1} and Fig.~\ref{fig:visual comparison 2} show the visualization results of different methods at sampling rates of 0.1 and 0.01, respectively. Notably, MB-RACS outperforms other methods in reconstructing detailed textures and sharper edges.

\subsubsection{Comparison of Sampling Rate Allocation}

Fig.~\ref{fig:visual and rate allocation comparison 1} and Fig.~\ref{fig:visual and rate allocation comparison 2} respectively display the reconstruction and sampling rate allocation results of MB-RACS and CASNet with two-stage rate-adaptive sampling and single-stage rate-adaptive sampling for an image from Set14 at a sampling rate of 0.04 and an image from Urban100 at a sampling rate of 0.25. In Fig.~\ref{fig:visual and rate allocation comparison 1}, it can be observed that the sampling rate allocation of MB-RACS is relatively uniform over the foreground objects, while CASNet's allocation is more focused on the tree to the right of the foreground and the rooftop of the house in front. Comparatively, MB-RACS's rate allocation may align more with human perception. In the region marked with a green box, where CASNet's sampling rate allocation tends to overlook, the reconstruction of MB-RACS exhibits richer textures. In Fig.~\ref{fig:visual and rate allocation comparison 2}, MB-RACS's rate allocation is relatively even across the foreground butterfly and the flower cluster on the left, whereas CASNet's allocation centers more on the butterfly. This could result in MB-RACS's reconstructed details being more enriched in the flower cluster region enclosed by the green box.

\subsubsection{Ablation Studies}
\label{sec:4:result and analysis:ablation studies}
Tab.~\ref{tab:experiments of ablation studies of average PSNR across datasets} shows the experiments of ablation studies on average PSNR across 7 test datasets under 5 sampling rates. Specifically, Tab.\ref{tab:experiments of ablation studies of average PSNR across datasets} (1) represents our single-stage rate-adaptive sampling model. Tab.\ref{tab:experiments of ablation studies of average PSNR across datasets} (2)-(4) are variants based on Tab.\ref{tab:experiments of ablation studies of average PSNR across datasets} (1) that are designed for ablation studies. Tab.~\ref{tab:experiments of ablation studies of average PSNR across datasets} (5)-(8) represent our multi-stage rate-adaptive sampling models with different numbers of sampling stages.

We trained a variant that replaces the rate-adaptive sampling module in our method with uniform sampling. The results are presented in Tab.~\ref{tab:experiments of ablation studies of average PSNR across datasets} (2). Compared to the uniform sampling variant, our single-stage measurement-bounds-based rate-adaptive sampling model achieves an average improvement of 1.89 dB. This validates the effectiveness of our proposed method for allocating sampling rates.

In Sec.~\ref{sec:3:single-stage measurement-bounds-based rate-adaptive sampling:sparsity threshold}, we propose a method for determining the overall sparsity ratio based on the overall sampling rate. To test the effectiveness of this method, we trained a variant with a fixed overall sparsity ratio of 0.15, the results of which are recorded in Tab.~\ref{tab:experiments of ablation studies of average PSNR across datasets} (3). Compared to this variant, our method achieves an average improvement of 0.72~dB. This confirms the effectiveness of adjusting the overall sparsity ratio based on the overall sampling rate.

We trained a variant without decoder-side cross-iteration skip connections, the results of which are shown in Tab.~\ref{tab:experiments of ablation studies of average PSNR across datasets} (4). Compared to this variant, our method achieves an average improvement of 0.09~dB. This confirms that decoder-side cross-iteration skip connections have a certain positive impact on performance.

\input{Section/experiments_and_results_section/tables/single-stage_table_avg_PSNR_across_datasets_ablation_expriment_results}

In the case of multi-stage rate-adaptive sampling, more sampling stages are expected to allocate the sampling rate more closely in proportion to the actual measurement bounds, thus potentially achieving higher performance. Tab.~\ref{tab:experiments of ablation studies of average PSNR across datasets} (5)-(8) presents the results of our multi-stage sampling models with 2, 5, 8, and 10 sampling stages, respectively. We find that increasing the number of sampling stages can boost performance; however, there is a diminishing return on this enhancement. Specifically, the model with 5 sampling stages is better than the 2-stage model by an average of 0.30 dB. However, going from 5 to 8 stages only gives a small improvement of 0.05 dB on average. Beyond 8 stages, the gains are minimal; upgrading to a 10-stage model results in only a slight improvement of 0.004 dB.

\input{Section/experiments_and_results_section/figures/multi-stage_sampling_figure_ablation_experiment_mean_PSNR_results}

\input{Section/experiments_and_results_section/figures/single-stage_sampling_figure_robustness_experiment_rate_of_gain_change_results_against_change_in_test_images}


In Sec.~\ref{sec:3:multi-stage rate-adaptive sampling strategy:sampling process:sampling in subsequent stages}, we introduce a distribution ratio optimization method, which is described in \eqref{eq:distribution ratio optimization}. To test the effectiveness of this method, we trained a variant without distribution ratio optimization. This variant uses the original distribution ratios \(\{p_i\}\) rather than the optimized \(\{q_i^{\star}\}\) for allocating the sampling rates. Fig.~\ref{fig:multi-stage sampling ablation studies} shows the average PSNR across seven test datasets and seven sampling rates for the proposed multi-stage rate-adaptive sampling method and this variant. Compared to the variant, our distribution ratio optimization method achieves an average improvement of 0.21~dB. This validates the effectiveness of the distribution ratio optimization method.

\subsubsection{Robustness Experiments for Measurement-Bounds-based Rate-Adaptive Sampling}
To investigate the robustness of the proposed measurement-bound-based rate-adaptive sampling to variations in brightness, scale, and rotation angle, we adjusted these parameters in our test images. We set the brightness and scale of the original test images to $1$, and then proportionally adjusted the brightness and scale of the test images. We also set the rotation angle of the original test images to $0^\circ$ , and during testing, rotated the images around their center. Moreover, we conducted experiments on the saliency-based rate-adaptive sampling proposed in CASNet \cite{9854112} as a reference. To evaluate the variability of performance gains under different parameter settings, we introduce a metric called the rate of gain change (RGC):
\begin{equation}
	\text{RGC} = \frac{\text{performance improvement after change}}{\text{performance improvement before change}} - 1
\end{equation}
For example, when the brightness is $0.5$, the PSNR gain achieved by a model using rate-adaptive sampling compared to one using uniform sampling is $1.1$ dB. When the brightness is $1$, the PSNR gain of the model using rate-adaptive sampling is $1$ dB. Then the rate of gain change at brightness $0.5$ is $\frac{1.1}{1} - 1 = 10\%$.

Fig.~\ref{fig:single-stage sampling robustness experiments rate of gain change results against change in test images} shows the rate of gain change in average PSNR across seven test datasets and seven sampling rates under varying brightness, scale, and rotation angles for MB-RACS with single-stage sampling and its variant with saliency-based rate-adaptive sampling, using the uniform sampling variant of MB-RACS as the baseline. For variations in brightness, our method outperforms the saliency-based rate-adaptive variant, with RGC metrics being $12.96\%$ and $34.42\%$ higher at brightness levels of $0.7$ and $1.4$, respectively. The differences in RGC at other brightness levels are relatively smaller. In terms of scale and rotation angle variations, our method consistently outperforms this variant, with an average advantage of $65.12\%$ for scale changes and $32.35\%$ for rotation angle changes. This suggests that the measurement-bounds-based rate-adaptive sampling demonstrates relatively better robustness in terms of performance gains compared to the saliency-based rate-adaptive sampling when subject to variations in brightness, scale, and rotation angle.

\input{Section/experiments_and_results_section/figures/single-stage_sampling_figure_robustness_experiment_psnrs_against_different_decoders}


Fig.~\ref{fig:single-stage sampling robustness experiments psnrs against different decoders} shows the average PSNR results of our single-stage measurement-bounds-based rate-adaptive sampling, single-stage saliency-based rate-adaptive sampling, and uniform sampling when combined with different decoders across six test datasets (excluding Urban100) and seven sampling rates. The Urban100 dataset was excluded because its relatively large image sizes caused the GPU memory requirements to surpass our current GPUs' capabilities when deploying the TransCS decoder. From the results, it is evident that our measurement-bounds-based rate-adaptive sampling method, in comparison with the uniform sampling combined with various decoders, shows a relatively stable improvement, exhibiting a performance increase with a standard deviation of $0.21$dB. On the other hand, the performance variability for saliency-based rate-adaptive sampling in relation to the uniform sampling has a standard deviation of $1.57$dB. Specifically, the performance of the saliency-based rate-adaptive sampling when combined with ISTA-Net$^{+}$ and OPINE-Net decoders appears less stable, registering performance declines of $3.62$dB and $0.07$dB, respectively, relative to the uniform sampling. These findings indicate that our measurement-bounds-based rate-adaptive sampling demonstrates better robustness to decoder variations than the saliency-based rate-adaptive sampling.

%% file: Section/experiments_and_results_section/figures/two-stage_figure_two_methods_visual_and_ratio_allocation_results_sampling_rate_0.04.tex
\begin{figure}
    \centering
    \newlength{\mylengthf}
    \setlength{\mylengthf}{\dimexpr 49.7\linewidth/100}
    \newlength{\mylengthg}
    \setlength{\mylengthg}{\dimexpr 49.1\linewidth/200}
    \begin{minipage}[t]{\mylengthf}
            \vspace{0pt}
		\centering
            \centerline{\footnotesize{Groud Truth}}
            \vspace{3.6pt}
		\includegraphics[width=\linewidth]{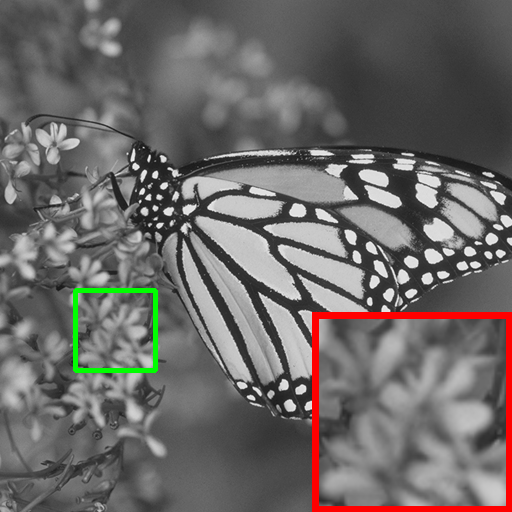}
  
            \vspace{-5pt}
            
		\centerline{\footnotesize{PSNR/SSIM}}	
            		
    \end{minipage}%
    \hfill
    \begin{minipage}[t]{\mylengthg}
            \vspace{0pt}
		\centering
            \centerline{\footnotesize{CASNet$_\text{ms;2}$}}
            \vspace{1.5pt}
		\includegraphics[width=\linewidth]{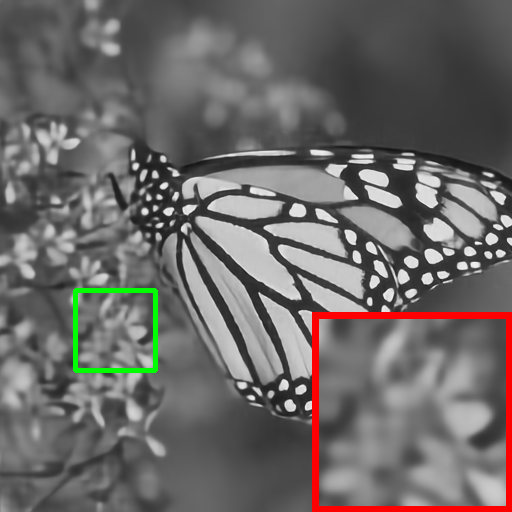}	

            \vspace{1.52pt}
  
            \includegraphics[width=\linewidth]{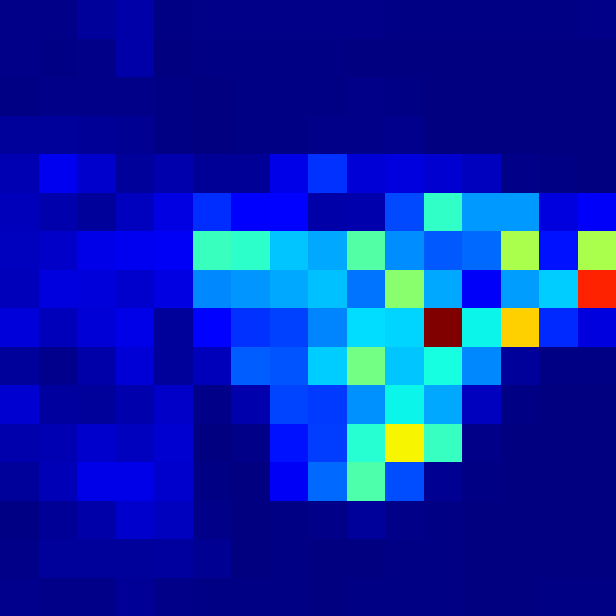}

            \vspace{-5pt}
            
		\centerline{\footnotesize{30.57/0.9077}}	
    \end{minipage}%
    \hfill
    \begin{minipage}[t]{\mylengthg}
            \vspace{0pt}
		\centering
            \centerline{\footnotesize{MB-RACS$_\text{ms;2}$}}
            \vspace{1.5pt}
		\includegraphics[width=\linewidth]{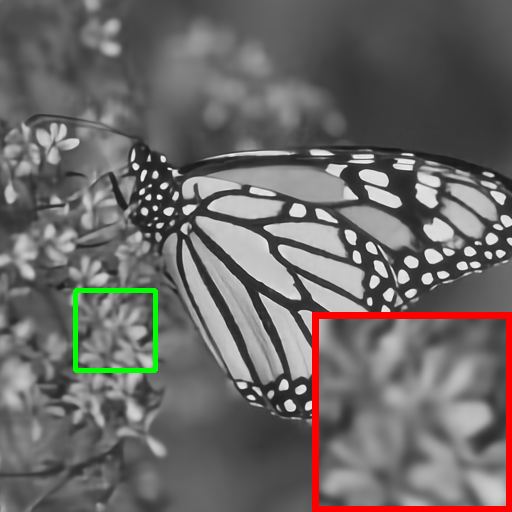}	
  
            \vspace{1.52pt}
            
            \includegraphics[width=\linewidth]{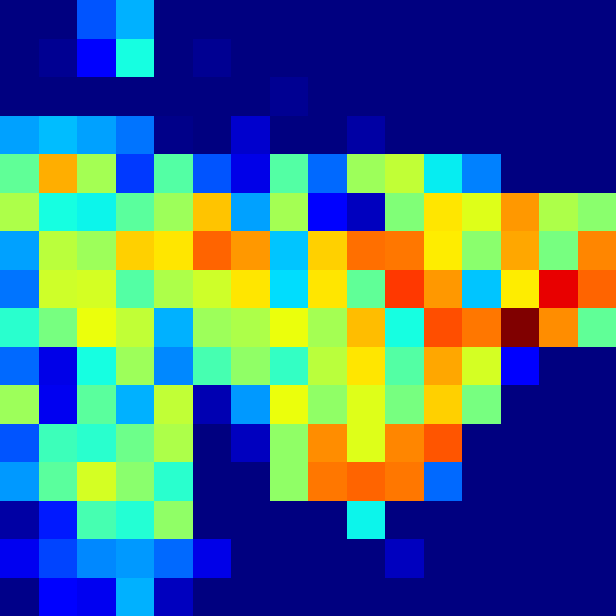}

            \vspace{-5pt}
            
		\centerline{\footnotesize{31.74/0.9262}}	
    \end{minipage}%
    
    \caption{Sampling rate allocation comparison between MB-RACS and CASNet with two-stage rate-adpative sampling on an Set14 image at 0.04 sampling rate.}
    \label{fig:visual and rate allocation comparison 1}
\end{figure}

%% file: Section/experiments_and_results_section/figures/single-stage_figure_two_methods_visual_and_ratio_allocation_results_sampling_rate_0.25.tex
\begin{figure}[t]
    \centering
    \newlength{\mylengthd}
    \setlength{\mylengthd}{\dimexpr 49.7\linewidth/100}
    \newlength{\mylengthe}
    \setlength{\mylengthe}{\dimexpr 49.1\linewidth/200}
    \begin{minipage}[t]{\mylengthd}
            \vspace{0pt}
		\centering
            \centerline{\footnotesize{Groud Truth}}
            \vspace{3.6pt}
		\includegraphics[width=\linewidth]{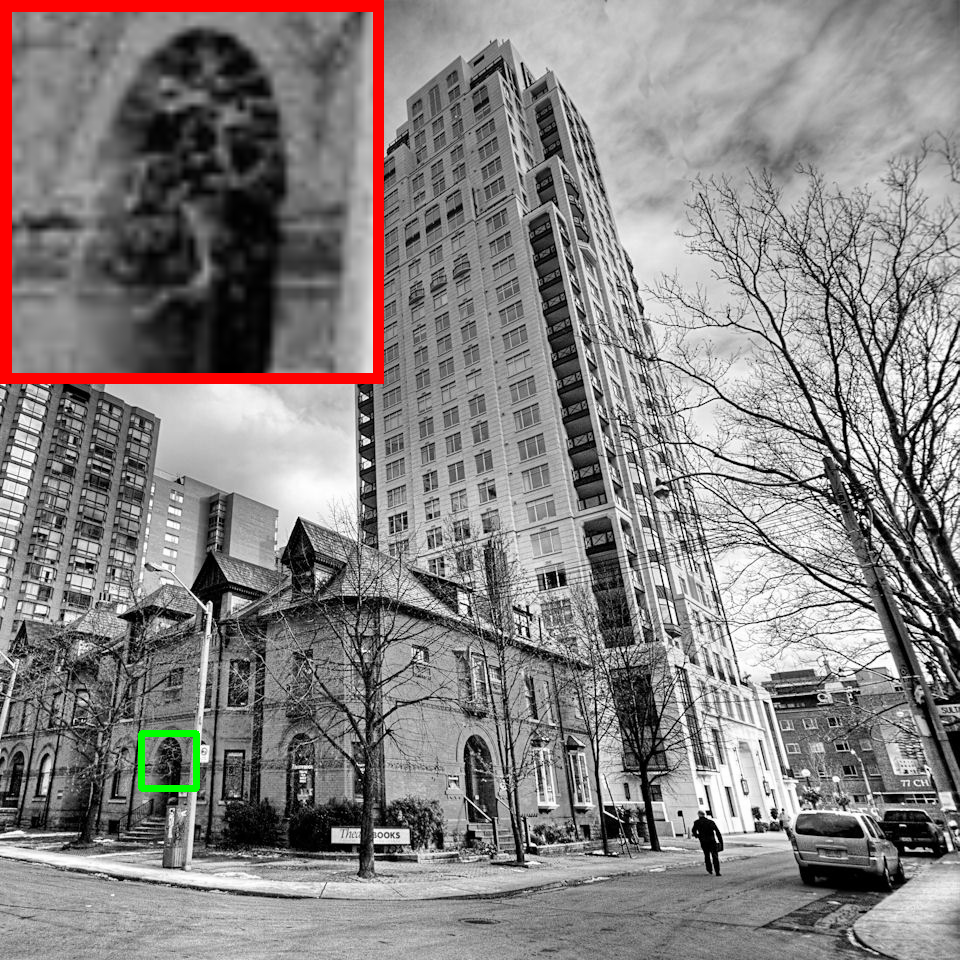}
  
            \vspace{-5pt}
            
		\centerline{\footnotesize{PSNR/SSIM}}	
            		
    \end{minipage}%
    \hfill
    \begin{minipage}[t]{\mylengthe}
            \vspace{0pt}
		\centering
            \centerline{\footnotesize{CASNet$_\text{ss}$}}
            \vspace{2.6pt}
		\includegraphics[width=\linewidth]{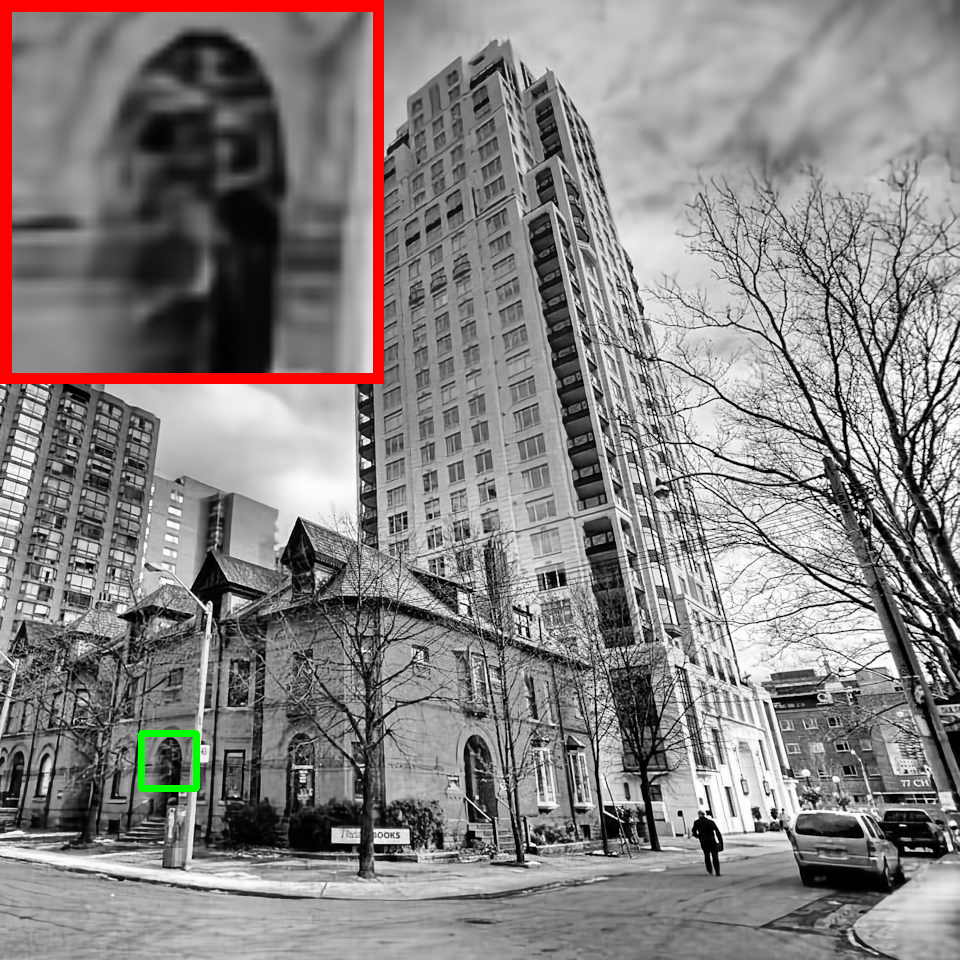}	

            \vspace{1.52pt}
  
            \includegraphics[width=\linewidth]{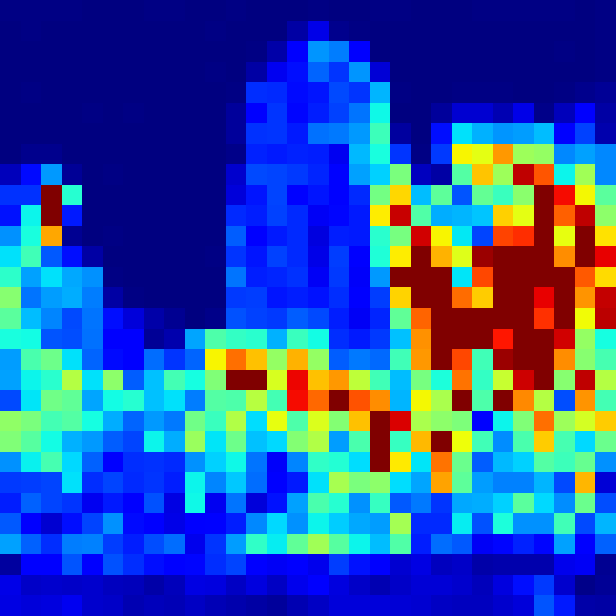}

            \vspace{-5pt}
            
		\centerline{\footnotesize{28.53/0.8932}}	
    \end{minipage}%
    \hfill
    \begin{minipage}[t]{\mylengthe}
            \vspace{0pt}
		\centering
            \centerline{\footnotesize{MB-RACS$_\text{ss}$}}
            \vspace{2.6pt}
		\includegraphics[width=\linewidth]{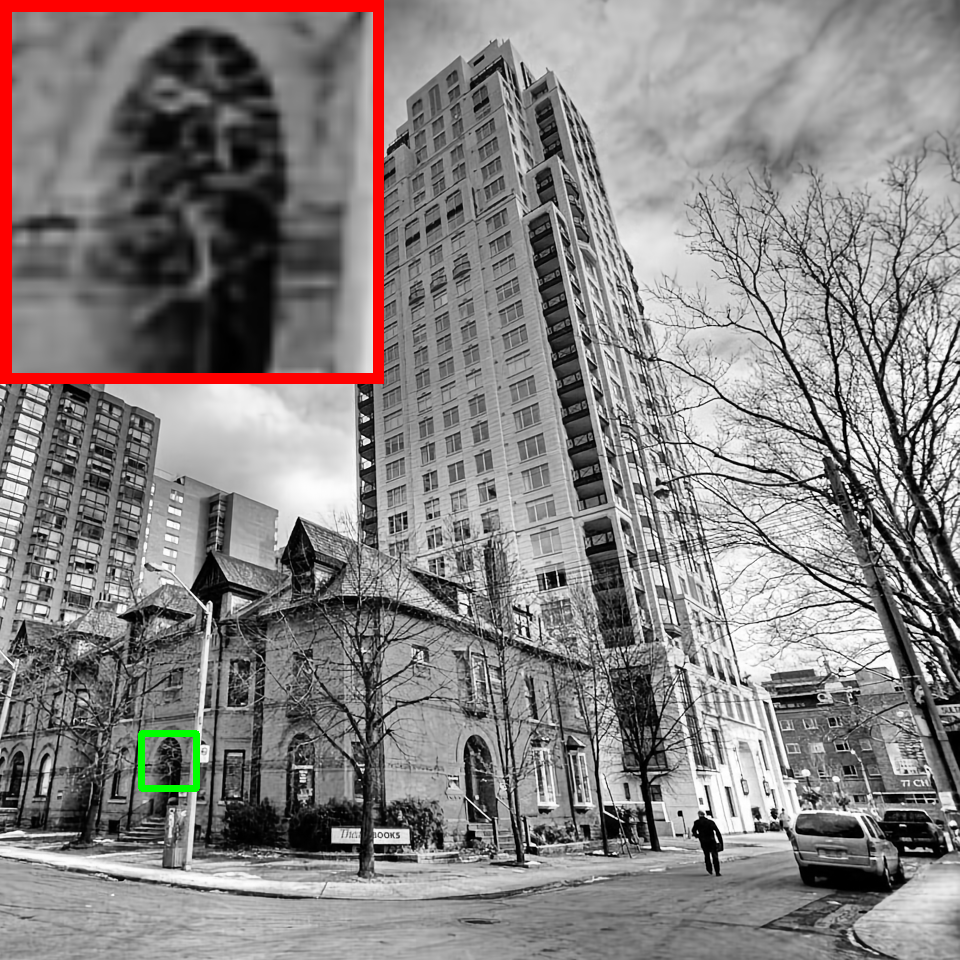}	
  
            \vspace{1.52pt}
            
            \includegraphics[width=\linewidth]{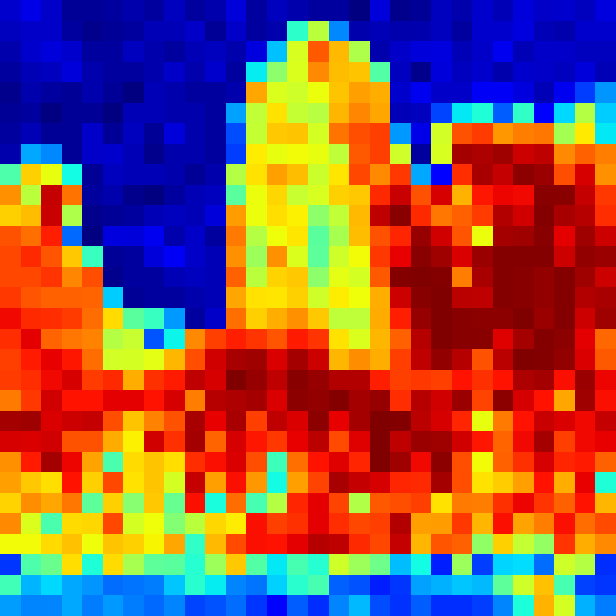}

            \vspace{-5pt}
            
		\centerline{\footnotesize{28.85/0.9269}}	
    \end{minipage}%
    
    \caption{Sampling rate allocation comparison between MB-RACS and CASNet with single-stage rate-adaptive sampling on an Urban100 image at 0.25 sampling rate.}
    \label{fig:visual and rate allocation comparison 2}
\end{figure}

%% file: Section/experiments_and_results_section/tables/single-stage_table_avg_PSNR_across_datasets_ablation_expriment_results.tex
\begin{table}
\centering
\caption{Experiments of ablation studies with 5 sampling rates using average PSNR across Set11, CBSD68, Set5, Set14, BSDS100, Urban100, and Kodak24 Datasets. }
\label{tab:experiments of ablation studies of average PSNR across datasets}
\resizebox{\linewidth}{!}{
\begin{tabular}{m{2.9cm} c c c c c}
\toprule
\multirow{2}{*}[-0.3em]{Setting} & \multicolumn{5}{c}{Sampling Ratio} \\ \cmidrule(lr){2-6} 
& \multicolumn{1}{c}{0.01} & \multicolumn{1}{c}{0.1} & \multicolumn{1}{c}{0.25} & \multicolumn{1}{c}{0.4} & \multicolumn{1}{c}{0.5} \\
\midrule
(1) 1-Stage Sampling & 23.49 & 30.85 & 35.49 & 39.01 & 41.31 \\ \midrule
(2) Uniform Sampling & 22.53 & 29.29 & 33.60 & 36.71 & 38.55 \\ 
\specialrule{0em}{1pt}{0pt}
(3) Fixed Sparsity Ratio & 22.82 & 30.04 & 34.75 & 38.33 & 40.63 \\ 
\specialrule{0em}{1pt}{0pt}
(4) w/o Skip Connections & 23.47 & 30.76 & 35.38 & 38.90 & 41.21 \\ \midrule
(5) 2-Stage Sampling & 22.98 & 30.39 & 34.97 & 38.43 & 40.55 \\ 
\specialrule{0em}{1pt}{0pt}
(6) 5-Stage Sampling & 23.13 & 30.65 & 35.29 & 38.77 & 40.98 \\ 
\specialrule{0em}{1pt}{0pt}
(7) 8-Stage Sampling & 23.15 & 30.68 & 35.33 & 38.85 & 41.06 \\ 
\specialrule{0em}{1pt}{0pt}
(8) 10-Stage Sampling & 23.16 & 30.68 & 35.34 & 38.85 & 41.06 \\
 \bottomrule
 \end{tabular}
} 
\end{table}

%% file: Section/experiments_and_results_section/figures/multi-stage_sampling_figure_ablation_experiment_mean_PSNR_results.tex
\begin{figure}[t]
\centering
\includegraphics[width=\linewidth]{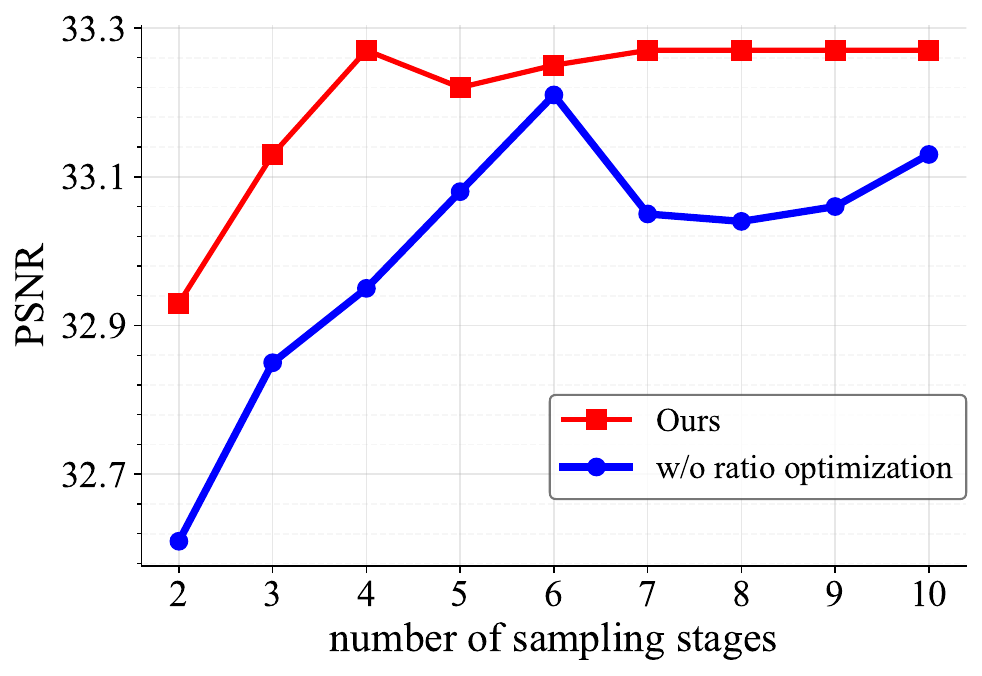}
\caption{Average PSNR curves, computed across seven test datasets and seven different sampling rates, are achieved by MB-RACS with multi-stage rate-adaptive sampling and its variant without ratio allocation optimization.}
\label{fig:multi-stage sampling ablation studies}
\end{figure}

%% file: Section/experiments_and_results_section/figures/single-stage_sampling_figure_robustness_experiment_rate_of_gain_change_results_against_change_in_test_images.tex
\begin{figure}[t]
\centering
\includegraphics[width=\linewidth]{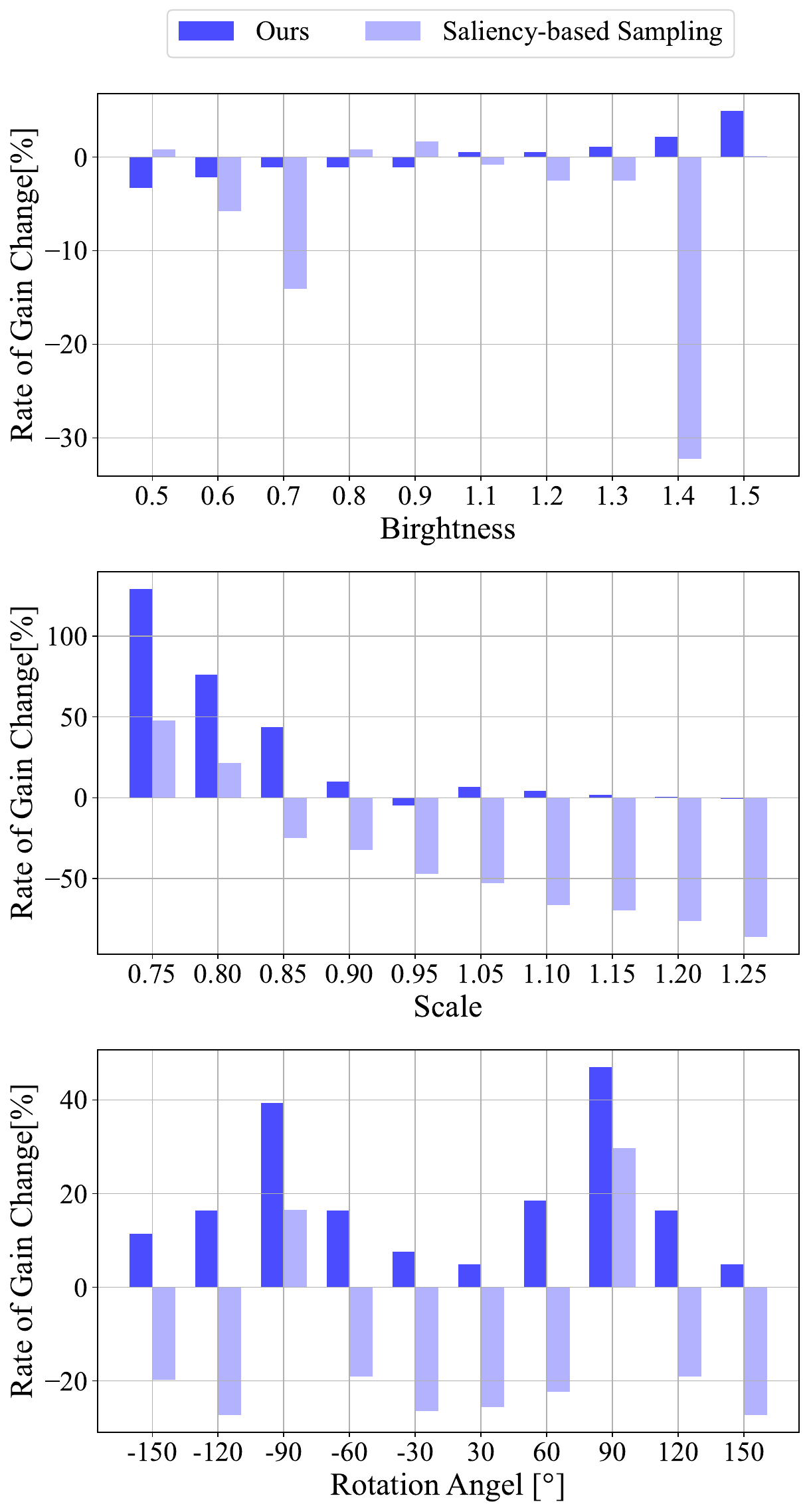}
\caption{Rate of gain change in average PSNR across seven test datasets and seven sampling rates achieved by MB-RACS with single-stage rate-adaptive sampling and its variant with saliency-based rate-adaptive sampling under different conditions of brightness, scale, and rotation angle in test images.}
\label{fig:single-stage sampling robustness experiments rate of gain change results against change in test images}
\end{figure}

%% file: Section/experiments_and_results_section/figures/single-stage_sampling_figure_robustness_experiment_psnrs_against_different_decoders.tex
\begin{figure}[t]
\centering
\includegraphics[width=\linewidth]{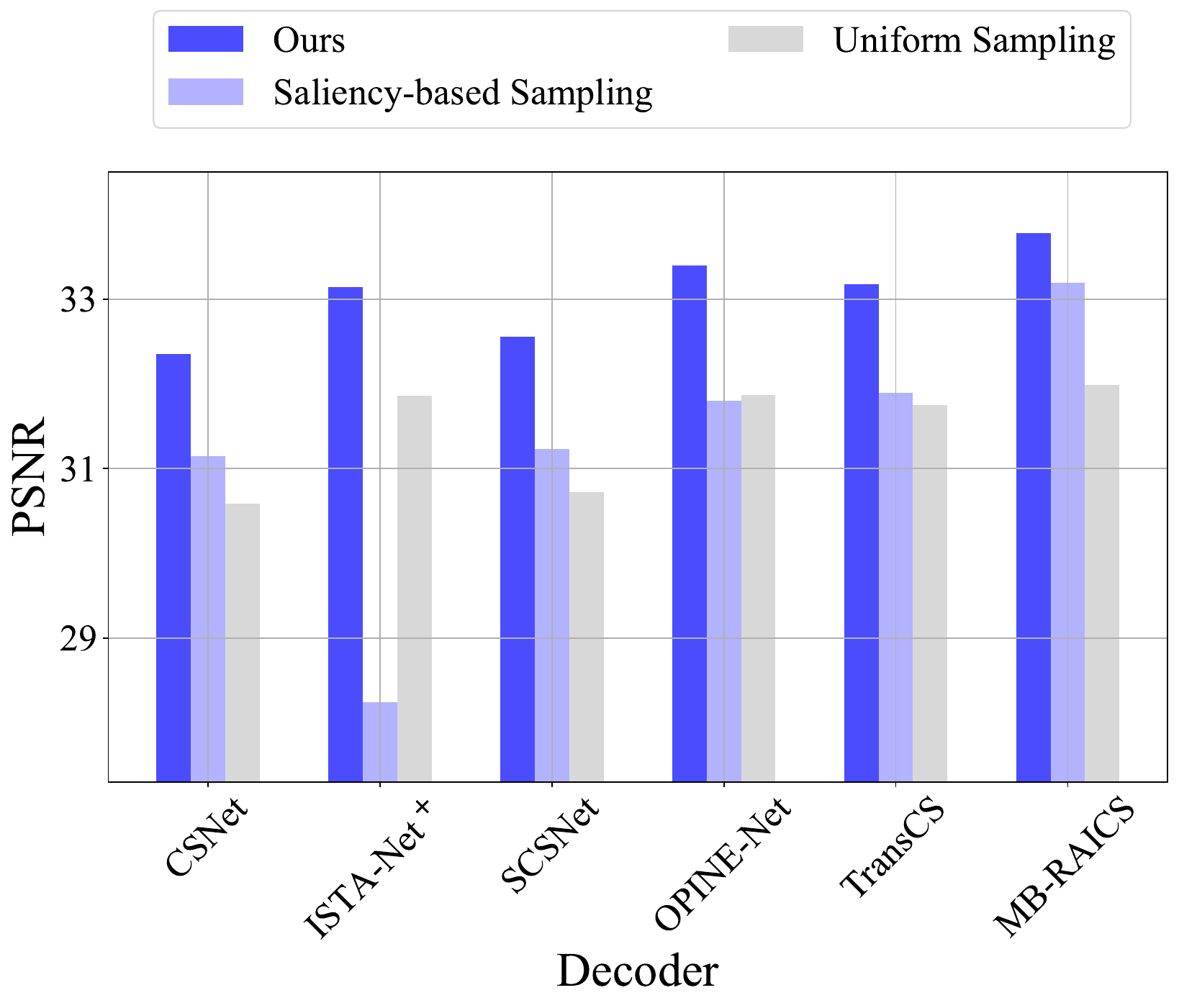}
\caption{Average PSNR results of our single-stage rate-adaptive sampling, saliency-based rate-adaptive sampling, and uniform sampling when combined with different decoders.}
\label{fig:single-stage sampling robustness experiments psnrs against different decoders}
\end{figure}

%% file: Section/conclusions.tex
\section{Conclusions}\label{sec:conclusions}
In this paper, we propose a measurement-bounds-based rate-adaptive sampling image compressed sensing method called MB-RACS, which adaptively allocates sampling rates to image blocks based on the theory of measurement bounds in traditional CS. In situations where it is not possible to obtain the statistical information of the possible real-world sources before sampling, we design an multi-stage rate-adaptive sampling strategy. MB-RACS employ priors from earlier sampling stages to determine the sampling rates for future stages. We also develop skip connections in the decoder to ensure thorough transmission of feature information. Extensive testing validates our method's effectiveness and confirms the strength of our rate-adaptive sampling based on measurement bounds.

%% file: Section/biography.tex
\begin{IEEEbiography}[{\includegraphics[width=1in,height=1.25in,clip,keepaspectratio]{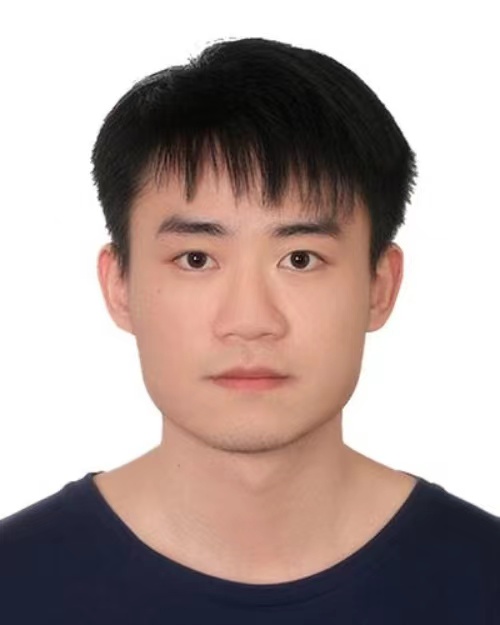}}]{Yujun Huang} received the bachelor's degree from Sun Yat-sen University, China, in 2020. He is currently a Ph.D. student of Department of Computer Science and Technology, Tsinghua University, under the supervisor of Prof. Shu-Tao Xia. His research interest generally includes Computer Vision and  Data Compression.
\end{IEEEbiography}

\begin{IEEEbiography}[{\includegraphics[width=1in,height=1.25in,clip,keepaspectratio]{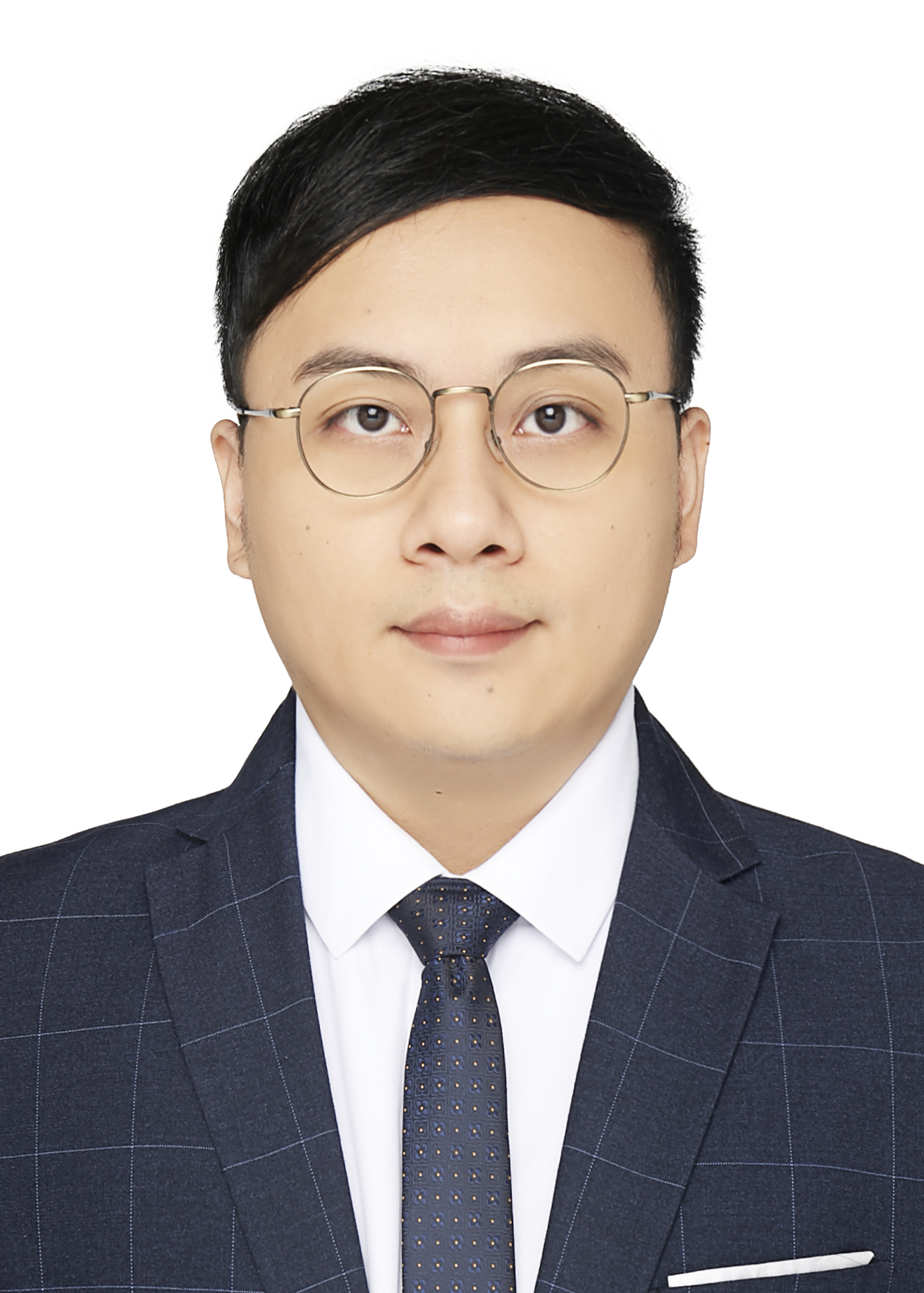}}]{Bin Chen} received the B.S. and M.S. degrees in mathematics from South China Normal University, Guangzhou, China, in 2014 and 2017, respectively, and Ph.D. degree from the Department of Computer Science and Technology, Tsinghua University, Beijing,  China, in 2021. From December 2019 to May 2020, he visited Department of Electrical and Computer Engineering, University of Waterloo, Canada. From May 2021 to November 2021, he was a Post-Doctoral Researcher with the Shenzhen International Graduate School, Tsinghua University, Shenzhen. Since December 2021, he has been an Assistant Professor with the Department of Computer Science and Technology, Harbin Institute of Technology, Shenzhen, China. He served as PC members/Program chairs for AAAI-21/22/23, IJCAI-21/22. His research interests include coding and information theory, machine learning, and deep learning.
\end{IEEEbiography}

\begin{IEEEbiography}[{\includegraphics[width=1in,height=1.25in,clip,keepaspectratio]{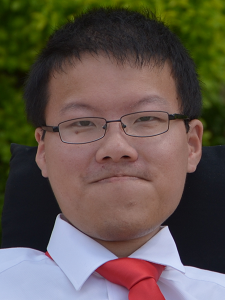}}]{Naiqi Li} received his Ph.D. degree in data science and information technology from Tsinghua University, China, in 2022. He is currently a postdoctoral research associate at  Tsinghua Shenzhen International Graduate School. 
His research interests include knowledge representation and reasoning, Gaussian processes, Bayesian inference, time series analysis and anomaly detection. His researches have been published in multiple top-tier venues, including NeurIPS, AAAI, IJCAI, ECAI, etc.
\end{IEEEbiography}

\begin{IEEEbiography}[{\includegraphics[width=1in,height=1.25in,clip,keepaspectratio]{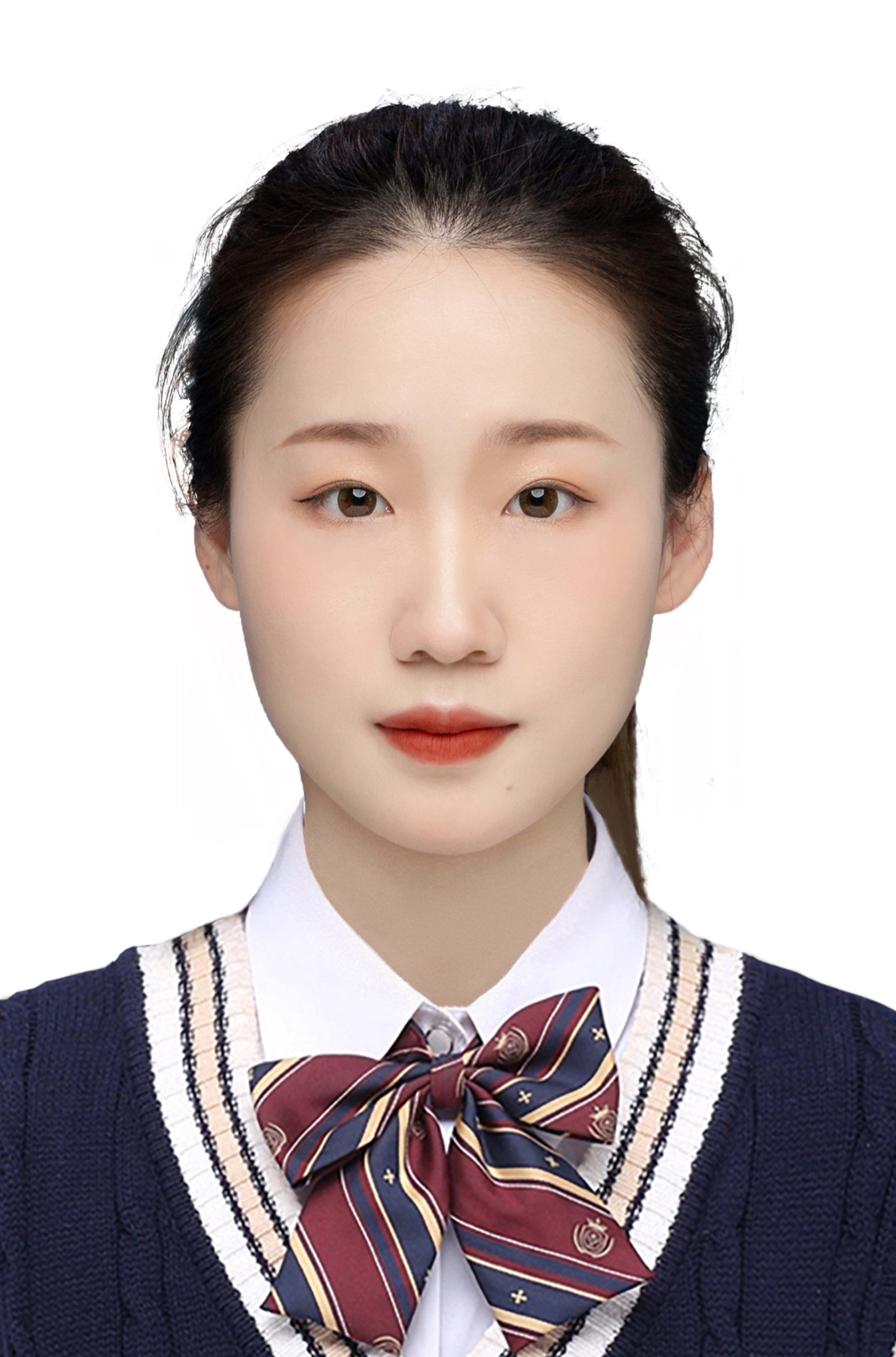}}]{Baoyi An} received the B.S. degree from the School of Computer Science and Technology, Nanjing Normal University, Nanjing, China, in 2017, and the PhD degree in computer science and technology with the School of Computer Science and Technology, University of Science and Technology of China, Hefei, China. She is currently a Senior Engineer with the Network Technology Lab, Central Research Institute, Huawei 2012 Labs. Her research interests include data compression, optimization theory and algorithms, mechanism design, blockchain, and privacy preservation.
\end{IEEEbiography}

\begin{IEEEbiography}[{\includegraphics[width=1in,height=1.25in,clip,keepaspectratio]{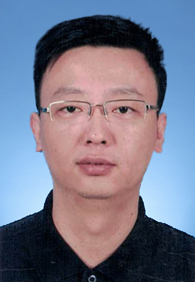}}]{Shu-Tao Xia} received the B.S. degree in mathematics and the Ph.D. degree in applied mathematics from Nankai University, Tianjin, China, in 1992 and 1997, respectively.

From March 1997 to April 1999, he was with the Research Group of Information Theory, Department of Mathematics, Nankai University. Since January 2004, he has been with the Shenzhen International Graduate School, Tsinghua University, Guangdong, China, where he is currently a Full Professor. His current research interests include coding and information theory, networking, and machine learning.
\end{IEEEbiography}

\begin{IEEEbiography}[{\includegraphics[width=1in,height=1.25in,clip,keepaspectratio]{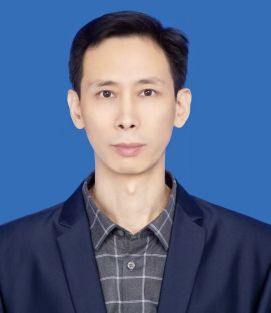}}]{Yaowei Wang} received the Ph.D. degree in computer science from the Graduate Uni- versity of Chinese Academy of Sciences, Beijing, China, in 2005. He is currently a tenured Associate Professor with Peng Cheng Laboratory, Shenzhen, China. He was with the Department of Electronics Engineering, Beijing Institute of Technology, from 2005 to 2019. He was a Professor with the Na- tional Engineering Laboratory for Video Technology Shenzhen (NELVT), Peking University Shenzhen Graduate School, Shenzhen, China, in 2019. From 2014 to 2015, he was an academic visitor with the Vision Lab of Queen Mary University of London. He is the author or co-author of more than 100 refereed journals and conference papers. His research interests include machine learning, multimedia content analysis, and understanding. He was the recipient of the second prize of the National Technology Invention in 2017 and the first prize of the CIE Technology Invention in 2015. His team was ranked as one of the best performers in the TRECVID CCD/SED tasks from 2009 to 2012 and in PETS 2012. He is a member of IEEE, CCF, CIE, and CSIG.
\end{IEEEbiography}

%% file: Section/appendix.tex
\clearpage
\appendices

\section{The KL Divergence-based Convex Optimization Problem}
\label{appendix:solution}

For notational convenience, we omit the superscript $t$ indicating the stage number in the following derivations.

Given allocation ratio $\textit{\textbf{p}}, \textit{\textbf{r}}\in \mathbb{R}^n$ ($\textbf{1}^{\transpose} \textit{\textbf{p}}=\textbf{1}^{\transpose} \textit{\textbf{r}}=1$, $\textit{\textbf{p}}\succeq\textbf{0}$, and $\textit{\textbf{r}}\succ\textbf{0}$), with $\alpha > 0$, $\beta > 0$, $\alpha + \beta = 1$, and position-specific upper bound $\textit{\textbf{a}}\in \mathbb{R}^n$ ($\textbf{0}\preceq \textit{\textbf{a}}\preceq\textbf{1}$), find the allocation ratio $\textbf{\textit{q}}^{\star}\in \mathbb{R}^n$ such that:

\begin{eqnarray}
\arg\min_{\textbf{\textit{q}}} \quad & -\sum_{i=1}^{n}p_i\log(\alpha q_i^{\star}+\beta r_i) \\
\text{subject to} \quad & \textbf{1}^{\transpose} \textbf{\textit{q}}^{\star}=1, \\ 
 & \textbf{0}\preceq \textbf{\textit{q}}^{\star}\preceq \textit{\textbf{a}}. 
\label{eq:convex optimization problem}
\end{eqnarray}

Formulate the constraints on $\textbf{\textit{q}}^{\star}$ as the KKT conditions:
\begin{align}
    & g_i^1(\textbf{\textit{q}}^{\star})=-q_i^{\star}\leq 0 \quad \text{for all} \quad i, \\
    & g_i^2(\textbf{\textit{q}}^{\star})=q_i^{\star}-a_i\leq 0 \quad \text{for all} \quad i, \\  
    & h(\textbf{\textit{q}}^{\star})=\textbf{1}^{\transpose} \textbf{\textit{q}}^{\star}-1=0.
\end{align}

And correspondingly introduce the Lagrange multipliers $\bm{\lambda}^{\star}, \bm{\pi}^{\star}\in \mathbb{R}^n$, and $\nu^{\star}$. Based on the KKT conditions, we can transform the problem of \eqref{eq:convex optimization problem} into the following problem:

\begin{align}
    \nabla_{\textbf{\textit{q}}^{\star}} \left(-\sum_{i=1}^{n}p_i \log(\alpha q_i^{\star}+\beta r_i) + \sum_{i=1}^n \lambda_i^{\star} g_i^1(\textbf{\textit{q}}^{\star})\right. \quad & \nonumber \\
    \quad \quad \quad \left. + \sum_{i=1}^n \pi_i^{\star} g_i^2(\textbf{\textit{q}}^{\star})+ \nu^{\star} h(\textbf{\textit{q}}^{\star})\right)=0 \quad \quad & \label{eq:convex optimization problem presented in KKT form}\\
    \text{subject to}  \quad  \lambda_i^{\star}g_i^1(\textbf{\textit{q}}^{\star})=0,\quad g_i^1(\textbf{\textit{q}}^{\star})\leq 0 \quad \text{for all} \quad i & \nonumber \\
     \quad \quad \pi_i^{\star}g_i^2(\textbf{\textit{q}}^{\star})=0,\quad
     g_i^2(\textbf{\textit{q}}^{\star})\leq0 \quad \text{for all} \quad i & \nonumber \\
     \quad \quad \bm{\lambda}^{\star}\succeq0,\quad
    \bm{\pi}^{\star}\succeq0,\quad h(\textbf{\textit{q}}^{\star})=0. \quad \quad \quad & \nonumber
\end{align}

\eqref{eq:convex optimization problem presented in KKT form} is equivalent to:
\begin{align}
    \frac{-\alpha p_i}{\alpha q_i^{\star}+\beta r_i}-
    \lambda_i^{\star}+\pi_i^{\star}+\nu^{\star}=0 \quad \text{for all} \quad i. \label{eq:convex optimization problem presented in simplified KKT form}
\end{align}

If $p_i = 0$, then $q_i^{\star} = 0$. Otherwise, the objective function in \eqref{eq:convex optimization problem} can be reduced by adding the value of $q_i$ to another $q_j$ corresponding to a $p_j$ that is greater than zero and setting $q_i = 0$. Therefore, we only consider the case where $p_i > 0$ in the following. 

If $\nu^{\star} \leq 0$, then according to the previous assumption that $p_i > 0$, $\pi_i^{\star} > 0$ is necessary for \eqref{eq:convex optimization problem presented in simplified KKT form} to hold. Also, since $\pi_i^{\star} g_i^2(\textbf{\textit{q}}^{\star}) = 0$, $q_i^{\star} = a_i$. As $a_i$ is the upper bound, it's reasonable to assume $\sum_{i=1}^n a_i > 1$. Thus $\sum_{i=1}^n q_i^{\star} = \sum_{i=1}^n a_i > 1$, which contradicts the condition, so we conclude that $\nu^{\star} > 0$.

If $p_i \geq \frac{\alpha a_i + \beta r_i}{\alpha} \nu^{\star}$, then we would observe that $\frac{-\alpha p_i}{\alpha q_i^{\star} + \beta r_i} + \nu^{\star} < 0$ when $q_i^{\star} < a_i$. \eqref{eq:convex optimization problem presented in simplified KKT form} can only hold if $\pi_i^{\star}>0$. Then, according to condition $\pi_i^{\star}g_i^2(\textbf{\textit{q}}^{\star})=0$, we can deduce that $q_i^{\star}=a_i$. This contradicts our previous assumption that $q_i^{\star} < a_i$, hence we conclude that $q_i^{\star} = a_i$.\\

If $\frac{\beta r_i}{\alpha}\nu^{\star}<p_i<\frac{\alpha a_i+\beta r_i}{\alpha}\nu^{\star}$, then $\frac{-\alpha p_i}{\alpha q_i^{\star}+\beta r_i}+\nu^{\star}>0$ will hold if $q_i^{\star}=a_i$. Then \eqref{eq:convex optimization problem presented in simplified KKT form} can only hold if $\lambda_i^{\star}>0$. According to condition $\lambda_i^{\star} g_i^1(\textbf{\textit{q}}^{\star})=0$, it can be deduced that $q_i^{\star}=0$ which contradicts the assumption above. So $q_i^{\star}<a_i$ and $\pi_i^{\star}=0$ must hold. If $q_i^{\star} = 0$, then $\frac{-\alpha p_i}{\alpha q_i^{\star}+\beta r_i}+\nu^{\star} < 0$, which makes \eqref{eq:convex optimization problem presented in simplified KKT form} untrue. Therefore, $q_i^{\star} > 0$ and $\lambda_i^{\star} = 0$. Lastly, from \eqref{eq:convex optimization problem presented in simplified KKT form}, we can deduce that $q_i^{\star} = \frac{p_i}{\nu^{\star}} - \frac{\beta r_i}{\alpha}$.

If $p_i \leq \frac{\beta r_i}{\alpha}\nu^{\star}$, then $q_i^{\star} > 0$ is not possible, because this would result in $\frac{-\alpha p_i}{\alpha q_i^{\star}+\beta r_i}+\nu^{\star} > 0$. From \eqref{eq:convex optimization problem presented in simplified KKT form}, we can see that this situation leads to $\lambda_i^{\star} > 0$. However, this contradicts the condition $\lambda_i^{\star}g_i^1(\textbf{\textit{q}}^{\star}) = 0$. Therefore, we conclude that $q_i^{\star} = 0$.

Thus we have
\begin{align}
q_i^{\star} = \begin{cases} a_i & p_{i} \geq \frac{\alpha a_i + \beta r_i}{\alpha} \nu^{\star} \\ \frac{p_i}{\nu^{\star}} - \frac{\beta r_i}{\alpha} & \frac{\beta r_i}{\alpha}\nu^{\star}<p_i<\frac{\alpha a_i+\beta r_i}{\alpha}\nu^{\star} \\ 0 & p_i \leq \frac{\beta r_i}{\alpha}\nu^{\star} \end{cases}.
 \label{eq:qi_solution}
\end{align}
Alternatively, a simplified representation is given by:
\begin{align}
q_i^{\star} = \min\left(\max\left(\frac{p_i}{\nu^{\star}} - \frac{\beta r_i}{\alpha}, 0\right), a_i\right). \label{eq:simplified_qi_solution}
\end{align}

Because $\sum_{i=1}^n q_i^{\star}= \sum_{i=1}^n \min\left(\max\left(\frac{p_i}{\nu^{\star}} - \frac{\beta r_i}{\alpha}, 0\right), a_i\right)=1$, $\nu^\star$ can be uniquely determined.

\section{The Inequality Relationship between the Intermediate Results of Newton's Method and the Root}
\label{appendix:lemma}

Similar to Appendix~\ref{appendix:solution}, we omit the superscript $t$ indicating the stage number in the following lemma and its proof.

\begin{lemma} \label{lemma:inequality relationship}
    Suppose the solution of the $\tau^{\text{th}}$ iteration in the Newton's method is $\mu^{(\tau)}$, then $\mu^{(\tau)} < \mu^{(\tau+1)}$ if and only if $\mu^{(\tau)} < \mu^{\star}$, and $\mu^{(\tau)} > \mu^{(\tau+1)}$ if and only if $\mu^{(\tau)} > \mu^{\star}$.
\end{lemma}

\begin{proof}
Suppose $\mu^{(\tau)} < \mu^{\star}$, if $\{N_\text{J}(\mu^{(\tau)}) | \text{J} = \text{l}, \text{c}, \text{u}\}$ and $\{N_\text{J}(\mu^{\star}) | \text{J} = \text{l}, \text{c}, \text{u}\}$ are completely equal, then:
\begin{align}
\mu^{(\tau+1)} &=\frac{1+\sum_{i\in N_\text{c}(\mu^{(\tau)})}\frac{\beta r_i}{\alpha}-\sum_{i\in N_\text{u}(\mu^{(\tau)})}a_i}{\sum_{i\in N_\text{c}(\mu^{(\tau)})}p_i} \nonumber \\
&=\frac{1+\sum_{i\in N_\text{c}(\mu^{\star})}\frac{\beta r_i}{\alpha}-\sum_{i\in N_\text{u}(\mu^{\star})}a_i}{\sum_{i\in N_\text{c}(\mu^{\star})}p_i} \nonumber \\
&=\mu^\star \nonumber.
\end{align}

Therefore, it can be deduced that $\mu^{(\tau)} < \mu^{\star} = \mu^{(\tau+1)}$. If $\{N_\text{J}(\mu^{(\tau)}) | \text{J} = l, c, u\}$ and $\{N_\text{J}(\mu^{\star}) | \text{J} = l, c, u\}$ are not completely equal, then $\mu^{(\tau)}$ and $\mu^{\star}$ are in the domains of different linear function segments. Therefore, $Q(\mu^{(\tau)}) < Q(\mu^{\star}) = 1$. Hence, we have:
\begin{align}
    Q(\mu^{(\tau)}) = \sum_{i\in N_\text{c}(\mu^{(\tau)})}(\mu^{(\tau)}p_i - \frac{\beta r_i}{\alpha}) + \sum_{i\in N_\text{u}(\mu^{(\tau)})}a_i < 1,  \nonumber 
\end{align}
Additionally, based on (24) in Section 3.2.2.3, we have:
\begin{align}
    \sum_{i\in N_\text{c}(\mu^{(\tau)})}(\mu^{(\tau+1)}p_i - \frac{\beta r_i}{\alpha}) + \sum_{i\in N_\text{u}(\mu^{(\tau)})}a_i = 1 \nonumber
\end{align}
After canceling out the equal terms in the above two expressions, we can obtain:
\begin{align}
    \mu^{(\tau)}\sum_{i\in N_\text{c}(\mu^{(\tau)})}p_i &< \mu^{(\tau+1)}\sum_{i\in N_\text{c}(\mu^{(\tau)})}p_i \nonumber \\
    \Rightarrow \quad \mu^{(\tau)} &< \mu^{(\tau+1)}. \nonumber
\end{align}
Thus, it is proven that if $\mu^{(\tau)} < \mu^\star$, then $\mu^{(\tau)} < \mu^{(\tau+1)}$. Using a similar method, it can be proven that if $\mu^{(\tau)} > \mu^\star$, then $\mu^{(\tau)} > \mu^{(\tau+1)}$. 

Additionally, based on the derivation of (24) in Section 3.2.2.3, it can be proven that if $\mu^{(\tau)} = \mu^\star$, then $\mu^{(\tau+1)} = \mu^\star$. In this case, $\mu^{(\tau)} = \mu^{(\tau+1)}$.

Based on the above propositions, we can draw the conclusion: $\mu^{(\tau)} < \mu^{(\tau+1)}$ if and only if $\mu^{(\tau)} < \mu^\star$. $\mu^{(\tau)} > \mu^{(\tau+1)}$ if and only if $\mu^{(\tau)} > \mu^\star$.
\end{proof}

%% file: main.bbl
\begin{thebibliography}{10}
\providecommand{\url}[1]{#1}
\csname url@samestyle\endcsname
\providecommand{\newblock}{\relax}
\providecommand{\bibinfo}[2]{#2}
\providecommand{\BIBentrySTDinterwordspacing}{\spaceskip=0pt\relax}
\providecommand{\BIBentryALTinterwordstretchfactor}{4}
\providecommand{\BIBentryALTinterwordspacing}{\spaceskip=\fontdimen2\font plus
\BIBentryALTinterwordstretchfactor\fontdimen3\font minus \fontdimen4\font\relax}
\providecommand{\BIBforeignlanguage}[2]{{%
\expandafter\ifx\csname l@#1\endcsname\relax
\typeout{** WARNING: IEEEtran.bst: No hyphenation pattern has been}%
\typeout{** loaded for the language `#1'. Using the pattern for}%
\typeout{** the default language instead.}%
\else
\language=\csname l@#1\endcsname
\fi
#2}}
\providecommand{\BIBdecl}{\relax}
\BIBdecl

\bibitem{1614066}
D.~Donoho, ``Compressed sensing,'' \emph{IEEE Transactions on Information Theory}, vol.~52, no.~4, pp. 1289--1306, 2006.

\bibitem{candes2006stable}
E.~J. Candes, J.~K. Romberg, and T.~Tao, ``Stable signal recovery from incomplete and inaccurate measurements,'' \emph{Communications on Pure and Applied Mathematics: A Journal Issued by the Courant Institute of Mathematical Sciences}, vol.~59, no.~8, pp. 1207--1223, 2006.

\bibitem{1542412}
E.~Candes and T.~Tao, ``Decoding by linear programming,'' \emph{IEEE Transactions on Information Theory}, vol.~51, no.~12, pp. 4203--4215, 2005.

\bibitem{4472246}
M.~Lustig, D.~L. Donoho, J.~M. Santos, and J.~M. Pauly, ``Compressed sensing mri,'' \emph{IEEE Signal Processing Magazine}, vol.~25, no.~2, pp. 72--82, 2008.

\bibitem{https://doi.org/10.1002/mrm.21391}
\BIBentryALTinterwordspacing
M.~Lustig, D.~Donoho, and J.~M. Pauly, ``Sparse mri: The application of compressed sensing for rapid mr imaging,'' \emph{Magnetic Resonance in Medicine}, vol.~58, no.~6, pp. 1182--1195, 2007. [Online]. Available: \url{https://onlinelibrary.wiley.com/doi/abs/10.1002/mrm.21391}
\BIBentrySTDinterwordspacing

\bibitem{8271999}
J.~Adler and O.~Öktem, ``Learned primal-dual reconstruction,'' \emph{IEEE Transactions on Medical Imaging}, vol.~37, no.~6, pp. 1322--1332, 2018.

\bibitem{yin2016compressive}
W.~Yin, X.~Fan, Y.~Shi, R.~Xiong, and D.~Zhao, ``Compressive sensing based soft video broadcast using spatial and temporal sparsity,'' \emph{Mobile Networks and Applications}, vol.~21, pp. 1002--1012, 2016.

\bibitem{6457441}
C.~Li, H.~Jiang, P.~Wilford, Y.~Zhang, and M.~Scheutzow, ``A new compressive video sensing framework for mobile broadcast,'' \emph{IEEE Transactions on Broadcasting}, vol.~59, no.~1, pp. 197--205, 2013.

\bibitem{265057}
\BIBentryALTinterwordspacing
Q.~Huang, S.~Sheng, X.~Chen, Y.~Bao, R.~Zhang, Y.~Xu, and G.~Zhang, ``Toward {Nearly-Zero-Error} sketching via compressive sensing,'' in \emph{18th USENIX Symposium on Networked Systems Design and Implementation (NSDI 21)}.\hskip 1em plus 0.5em minus 0.4em\relax USENIX Association, Apr. 2021, pp. 1027--1044. [Online]. Available: \url{https://www.usenix.org/conference/nsdi21/presentation/huang}
\BIBentrySTDinterwordspacing

\bibitem{5414429}
S.~Mun and J.~E. Fowler, ``Block compressed sensing of images using directional transforms,'' in \emph{2009 16th IEEE International Conference on Image Processing (ICIP)}, 2009, pp. 3021--3024.

\bibitem{li2013efficient}
C.~Li, W.~Yin, H.~Jiang, and Y.~Zhang, ``An efficient augmented lagrangian method with applications to total variation minimization,'' \emph{Computational Optimization and Applications}, vol.~56, pp. 507--530, 2013.

\bibitem{6341094}
J.~Zhang, D.~Zhao, C.~Zhao, R.~Xiong, S.~Ma, and W.~Gao, ``Image compressive sensing recovery via collaborative sparsity,'' \emph{IEEE Journal on Emerging and Selected Topics in Circuits and Systems}, vol.~2, no.~3, pp. 380--391, 2012.

\bibitem{6814320}
J.~Zhang, D.~Zhao, and W.~Gao, ``Group-based sparse representation for image restoration,'' \emph{IEEE Transactions on Image Processing}, vol.~23, no.~8, pp. 3336--3351, 2014.

\bibitem{8765626}
W.~Shi, F.~Jiang, S.~Liu, and D.~Zhao, \emph{IEEE Transactions on Image Processing}, vol.~29, pp. 375--388, 2020.

\bibitem{Zhang_2018_CVPR}
J.~Zhang and B.~Ghanem, ``Ista-net: Interpretable optimization-inspired deep network for image compressive sensing,'' in \emph{Proceedings of the IEEE Conference on Computer Vision and Pattern Recognition (CVPR)}, June 2018.

\bibitem{9298950}
Z.~Zhang, Y.~Liu, J.~Liu, F.~Wen, and C.~Zhu, ``Amp-net: Denoising-based deep unfolding for compressive image sensing,'' \emph{IEEE Transactions on Image Processing}, vol.~30, pp. 1487--1500, 2021.

\bibitem{9019857}
J.~Zhang, C.~Zhao, and W.~Gao, ``Optimization-inspired compact deep compressive sensing,'' \emph{IEEE Journal of Selected Topics in Signal Processing}, vol.~14, no.~4, pp. 765--774, 2020.

\bibitem{5585813}
Y.~Yu, B.~Wang, and L.~Zhang, ``Saliency-based compressive sampling for image signals,'' \emph{IEEE Signal Processing Letters}, vol.~17, no.~11, pp. 973--976, 2010.

\bibitem{9159912}
S.~Zhou, Y.~He, Y.~Liu, C.~Li, and J.~Zhang, ``Multi-channel deep networks for block-based image compressive sensing,'' \emph{IEEE Transactions on Multimedia}, vol.~23, pp. 2627--2640, 2021.

\bibitem{9854112}
B.~Chen and J.~Zhang, ``Content-aware scalable deep compressed sensing,'' \emph{IEEE Transactions on Image Processing}, vol.~31, pp. 5412--5426, 2022.

\bibitem{Shi_2019_CVPR}
W.~Shi, F.~Jiang, S.~Liu, and D.~Zhao, ``Scalable convolutional neural network for image compressed sensing,'' in \emph{Proceedings of the IEEE/CVF Conference on Computer Vision and Pattern Recognition (CVPR)}, June 2019.

\bibitem{9467810}
D.~You, J.~Zhang, J.~Xie, B.~Chen, and S.~Ma, ``Coast: Controllable arbitrary-sampling network for compressive sensing,'' \emph{IEEE Transactions on Image Processing}, vol.~30, pp. 6066--6080, 2021.

\bibitem{doi:10.1137/S003614450037906X}
\BIBentryALTinterwordspacing
S.~S. Chen, D.~L. Donoho, and M.~A. Saunders, ``Atomic decomposition by basis pursuit,'' \emph{SIAM Review}, vol.~43, no.~1, pp. 129--159, 2001. [Online]. Available: \url{https://doi.org/10.1137/S003614450037906X}
\BIBentrySTDinterwordspacing

\bibitem{258082}
S.~Mallat and Z.~Zhang, ``Matching pursuits with time-frequency dictionaries,'' \emph{IEEE Transactions on Signal Processing}, vol.~41, no.~12, pp. 3397--3415, 1993.

\bibitem{f957882f5b1e4f3ba0450468093903f1}
I.~Daubechies, M.~Defrise, and C.~{De Mol}, ``\BIBforeignlanguage{English}{An iterative thresholding algorithm for linear inverse problems with a sparsity constraint},'' \emph{\BIBforeignlanguage{English}{Unknown Journal}}, 2004, communications in Pure and Applied Math., 57, no 11, 1413-1457 (2004).

\bibitem{7923735}
J.~Mu, R.~Xiong, X.~Zhang, and S.~Ma, ``Band-wise adaptive sparsity regularization for quantized compressed sensing exploiting nonlocal similarity,'' in \emph{2017 Data Compression Conference (DCC)}, 2017, pp. 452--452.

\bibitem{Kulkarni_2016_CVPR}
K.~Kulkarni, S.~Lohit, P.~Turaga, R.~Kerviche, and A.~Ashok, ``Reconnet: Non-iterative reconstruction of images from compressively sensed measurements,'' in \emph{Proceedings of the IEEE Conference on Computer Vision and Pattern Recognition (CVPR)}, June 2016.

\bibitem{doi:10.1137/080716542}
\BIBentryALTinterwordspacing
A.~Beck and M.~Teboulle, ``A fast iterative shrinkage-thresholding algorithm for linear inverse problems,'' \emph{SIAM Journal on Imaging Sciences}, vol.~2, no.~1, pp. 183--202, 2009. [Online]. Available: \url{https://doi.org/10.1137/080716542}
\BIBentrySTDinterwordspacing

\bibitem{9934025}
M.~Shen, H.~Gan, C.~Ning, Y.~Hua, and T.~Zhang, ``Transcs: A transformer-based hybrid architecture for image compressed sensing,'' \emph{IEEE Transactions on Image Processing}, vol.~31, pp. 6991--7005, 2022.

\bibitem{NIPS2017_3f5ee243}
\BIBentryALTinterwordspacing
A.~Vaswani, N.~Shazeer, N.~Parmar, J.~Uszkoreit, L.~Jones, A.~N. Gomez, L.~u. Kaiser, and I.~Polosukhin, ``Attention is all you need,'' in \emph{Advances in Neural Information Processing Systems}, I.~Guyon, U.~V. Luxburg, S.~Bengio, H.~Wallach, R.~Fergus, S.~Vishwanathan, and R.~Garnett, Eds., vol.~30.\hskip 1em plus 0.5em minus 0.4em\relax Curran Associates, Inc., 2017. [Online]. Available: \url{https://proceedings.neurips.cc/paper_files/paper/2017/file/3f5ee243547dee91fbd053c1c4a845aa-Paper.pdf}
\BIBentrySTDinterwordspacing

\bibitem{BMVC.26.135}
M.~Bevilacqua, A.~Roumy, C.~Guillemot, and M.~line Alberi~Morel, ``Low-complexity single-image super-resolution based on nonnegative neighbor embedding,'' in \emph{Proceedings of the British Machine Vision Conference}.\hskip 1em plus 0.5em minus 0.4em\relax BMVA Press, 2012, pp. 135.1--135.10.

\bibitem{4472247}
M.~F. Duarte, M.~A. Davenport, D.~Takhar, J.~N. Laska, T.~Sun, K.~F. Kelly, and R.~G. Baraniuk, ``Single-pixel imaging via compressive sampling,'' \emph{IEEE Signal Processing Magazine}, vol.~25, no.~2, pp. 83--91, 2008.

\bibitem{He_2016_CVPR}
K.~He, X.~Zhang, S.~Ren, and J.~Sun, ``Deep residual learning for image recognition,'' in \emph{Proceedings of the IEEE Conference on Computer Vision and Pattern Recognition (CVPR)}, June 2016.

\bibitem{huang2017densely}
G.~Huang, Z.~Liu, L.~Van Der~Maaten, and K.~Q. Weinberger, ``Densely connected convolutional networks,'' in \emph{Proceedings of the IEEE conference on computer vision and pattern recognition}, 2017, pp. 4700--4708.

\bibitem{5557884}
P.~Arbeláez, M.~Maire, C.~Fowlkes, and J.~Malik, ``Contour detection and hierarchical image segmentation,'' \emph{IEEE Transactions on Pattern Analysis and Machine Intelligence}, vol.~33, no.~5, pp. 898--916, 2011.

\bibitem{937655}
D.~Martin, C.~Fowlkes, D.~Tal, and J.~Malik, ``A database of human segmented natural images and its application to evaluating segmentation algorithms and measuring ecological statistics,'' in \emph{Proceedings Eighth IEEE International Conference on Computer Vision. ICCV 2001}, vol.~2, 2001, pp. 416--423 vol.2.

\bibitem{bevilacqua2012low}
M.~Bevilacqua, A.~Roumy, C.~Guillemot, and M.~L. Alberi-Morel, ``Low-complexity single-image super-resolution based on nonnegative neighbor embedding,'' 2012.

\bibitem{zeyde2012single}
R.~Zeyde, M.~Elad, and M.~Protter, ``On single image scale-up using sparse-representations,'' in \emph{Curves and Surfaces: 7th International Conference, Avignon, France, June 24-30, 2010, Revised Selected Papers 7}.\hskip 1em plus 0.5em minus 0.4em\relax Springer, 2012, pp. 711--730.

\bibitem{huang2015single}
J.-B. Huang, A.~Singh, and N.~Ahuja, ``Single image super-resolution from transformed self-exemplars,'' in \emph{Proceedings of the IEEE conference on computer vision and pattern recognition}, 2015, pp. 5197--5206.

\bibitem{franzen1999kodak}
R.~Franzen, ``Kodak lossless true color image suite,'' \emph{source: http://r0k. us/graphics/kodak}, vol.~4, no.~2, p.~9, 1999.

\bibitem{wang2003multiscale}
Z.~Wang, E.~P. Simoncelli, and A.~C. Bovik, ``Multiscale structural similarity for image quality assessment,'' in \emph{The Thrity-Seventh Asilomar Conference on Signals, Systems \& Computers, 2003}, vol.~2.\hskip 1em plus 0.5em minus 0.4em\relax Ieee, 2003, pp. 1398--1402.

\bibitem{9767634}
Z.~Zheng, W.~Dai, D.~Xue, C.~Li, J.~Zou, and H.~Xiong, ``Hybrid ista: Unfolding ista with convergence guarantees using free-form deep neural networks,'' \emph{IEEE Transactions on Pattern Analysis and Machine Intelligence}, vol.~45, no.~3, pp. 3226--3244, 2023.

\end{thebibliography}
